\documentclass[11pt]{article}

\usepackage[final]{acl}
\usepackage{times}
\usepackage{latexsym}
\usepackage[T1]{fontenc}
\usepackage[utf8]{inputenc}
\usepackage{courier}

\usepackage{microtype}
\usepackage{graphicx}
\usepackage{subfigure}
\usepackage{booktabs} % for professional tables

\usepackage{enumitem}
\usepackage[most]{tcolorbox}
\tcbuselibrary{listings, breakable}
\newtcolorbox{promptbox}[2][]{%
  enhanced,
  breakable,
  colback=gray!5,
  colframe=gray!40,
  coltitle=black,
  fonttitle=\bfseries,
  attach boxed title to top left={yshift=-2mm, xshift=2mm},
  boxed title style={colback=white, colframe=gray!40, sharp corners},
  title={#2},
  sharp corners,
  boxrule=0.8pt,
  listing only,         
  listing options={
    basicstyle=\small\ttfamily,
    breaklines=true,
    columns=fullflexible
  },
  #1
}
\usepackage{amsmath}
\usepackage{amssymb}
\usepackage{mathtools}
\usepackage{amsthm}

\usepackage[capitalize,noabbrev]{cleveref}

\usepackage{booktabs}
\usepackage{tabularx}
\usepackage[table]{xcolor}
\usepackage{colortbl}
\usepackage{caption}
\usepackage{placeins}
\usepackage{colortbl}
\usepackage{booktabs}
\usepackage{graphicx}
\newcommand{\num}[1]{#1}
\usepackage{pgf}
\usepackage{pgfplots}
\pgfplotsset{compat=1.18}
\usepgfplotslibrary{groupplots}
\usepackage{booktabs}
\usepackage{multirow}
\usepackage{array} % for column control
\usepackage{xurl}
\usepackage{longtable}

\definecolor{HeatBlue}{RGB}{33,113,181}
\definecolor{AgentPink}{RGB}{242,83,177}
\definecolor{AgentTeal}{RGB}{35,191,184}
\definecolor{AblatePurple}{RGB}{121,109,224}
\definecolor{AblateGold}{RGB}{231,160,68}

\newcommand{\heat}[3]{%
  \pgfmathsetmacro{\v}{#1}%
  \pgfmathsetmacro{\minv}{#2}%
  \pgfmathsetmacro{\maxv}{#3}%
  \pgfmathsetmacro{\t}{(\v-\minv)/max(\maxv-\minv,1e-9)}%
  \pgfmathsetmacro{\p}{min(max(8 + 57*\t, 8), 65)}%
  \edef\applycellcolor{\noexpand\cellcolor{HeatBlue!\p}}%
  \applycellcolor
  \num{#1}%
}

\theoremstyle{plain}

\theoremstyle{definition}

\theoremstyle{remark}

\usepackage[textsize=tiny]{todonotes}

\title{AgentFold: Closed-Loop Agentic Search for Protein Folding Model Design}

\author{
\textbf{Mingquan Liu}$^{1}$\thanks{Equal contribution.} \quad
\textbf{Jiangyu Chen}$^{2}$\footnotemark[1] \quad
\textbf{Hanqun Cao}$^{3}$\footnotemark[1] \quad
\textbf{Xujun Zhang}$^{7}$ \quad
\textbf{Pengsen Ma}$^{1}$ \\
\textbf{Xiangru Tang}$^{5}$ \quad
\textbf{Shuting Jin}$^{4}$ \quad
\textbf{Annie Zheng}$^{6}$ \quad
\textbf{Zhuo Yang}$^{8}$ \quad
\textbf{Tianfan Fu}$^{2}$\thanks{Corresponding authors.} \\
\textbf{Fang Wu}$^{9}$\footnotemark[2] \quad
\textbf{Xiangxiang Zeng}$^{1}$\footnotemark[2] \\
\normalfont
$^{1}$State Key Lab. of Chemo \& Biosensing,
Coll. of Comp. Sci. \& Electron. Eng.,
Hunan University \\
$^{2}$State Key Laboratory for Novel Software Technology,
Sch. of Comput. Sci., Nanjing University \\
$^{3}$The Chinese University of Hong Kong \quad
$^{4}$Wuhan University of Science and Technology \\
$^{5}$Yale University \quad
$^{6}$South China Normal University Affiliated High School  \\
$^{7}$Zhejiang University \quad
$^{8}$Southeast University  \quad
$^{9}$Stanford University \\
\texttt{xzeng@hnu.edu.cn}
}

\begin{document}
\maketitle

\begin{abstract}
Scientific LLM agents have shown promise in literature reasoning, tool use, and experiment planning, but it remains unclear whether they can autonomously improve large, tightly coupled scientific ML systems through executable code changes and expensive validation. We study this question in protein folding, where progress requires coordinated architectural edits, multi-objective evaluation, and domain-aware interpretation. We present \textbf{AgentFold}, a multi-agent framework that formulates folding-model development as closed-loop search over executable code variants. Starting from ESMFold, AgentFold proposes hypotheses, implements and debugs code-level modifications, evaluates variants, analyzes outcomes, and stores both successful and failed interventions in structured memory; an MCTS-style policy allocates compute across high-scoring branches. On an engineering-scale folding codebase (\(>\textbf{2{,}000}\) LOC), AgentFold explores \({\sim}\textbf{80}\) variants using \({\sim}\textbf{5{,}000}\) GPU-hours and \({\sim}\textbf{170M}\) LLM tokens. At matched budget, AgentFold improves best lDDT by \(\textbf{7.5\%}\) over independent Codex proposals and beats random control. Beyond model improvement, the intervention traces reveal recurring empirical design patterns: stable gains tend to arise from early soft learnable priors and gated refinement, whereas direct geometric perturbations and geometry-conditioned feedback often destabilize training. The code and experimental resources are publicly available at
\url{https://github.com/lmqfly/AgentFold}.

\end{abstract}

\section{Introduction}

Scientific agents increasingly combine large language models (LLMs) with literature analysis, hypothesis generation, tool use, and experimental planning~\cite{lu2024ai,tang2025cellforge,Hsu2024CHIMELH,Qi2023LargeLM,Fallahpour2025BioReasonIM,Huang2025AutomatedHV,Hao2025PerTurboAgentAS,Huang2025BiomniAG,Wang2026SpatialAgentAA,Jin2026STELLATA}. Execution-grounded benchmarks separately show that iterative machine-learning experimentation and scientific-code development remain difficult even when outcomes can be checked automatically~\cite{Tian2024SciCodeAR,Huang2023MLAgentBenchEL,Huang2024DACodeAD,Chan2024MLEbenchEM,Edwards2025RExBenchCC}. In scientific ML, a plausible proposal is insufficient: the system must implement the change in a coupled codebase, recover from failures, and compare expensive, noisy, multi-objective experiments.

We study this question in protein folding, where architectural changes are executable interventions in a tightly coupled scientific ML system. Folding models combine sequence and pair representations, geometric refinement, recycling, and structure losses, while evaluation spans both local and global structural metrics. This setting provides a suitable testbed for assessing whether LLM agents are capable of closed-loop scientific model development beyond code generation assistance.

We introduce \textbf{AgentFold}, a multi-agent framework for code-level search over folding-model variants. Starting from a compact ESMFold-derived substrate~\cite{Lin2022EvolutionaryscalePO}, AgentFold executes a propose--implement--evaluate loop: it retrieves evidence from a folding-model zoo and a structured memory, proposes architectural or algorithmic edits, applies and debugs code changes, evaluates executable variants, and records both successful and failed interventions. Failed or low-performing variants are retained as structured evidence, allowing later proposals to avoid repeated failure modes and supporting post-hoc comparison among related edits. We use the compact substrate to enable repeated training and evaluation while preserving the coupled structure-module setting that makes folding-model design nontrivial.

To allocate compute over long-horizon exploration, AgentFold uses an MCTS-style tree controller over concrete code snapshots. Each node represents an executable implementation, while expansions are prioritized using standard folding metrics and a normalized search utility. On an engineering-scale codebase (>2{,}000 LOC), AgentFold explores roughly 80 variants using approximately 5,000 GPU-hours and 170M LLM tokens. At a matched evaluation budget on the CAMEO2022 \cite{Haas2017ContinuousAM} development benchmark, AgentFold achieves 7.5\% higher best lDDT than an independent Codex-proposal baseline and also outperforms a random-search controller. The strongest variants obtain these improvements with only modest parameter overhead. Analysis of both successful and failed interventions further reveals descriptive regularities: stable improvements frequently co-occur with early soft learnable priors and gated refinement, whereas direct geometric perturbations and geometry-conditioned feedback are often associated with training instability.

\paragraph{Contributions.}
Our contributions are:
\begin{itemize}[leftmargin=*]
    \item \textbf{Closed-loop folding model search.} We present AgentFold, a multi-agent framework that formulates folding-model development as propose--implement--debug--evaluate cycles over executable code variants rather than limiting the search to textual hypotheses.
    
    \item \textbf{Engineering-scale matched-budget evaluation.} Starting from a compact ESMFold-derived substrate, AgentFold evaluates roughly 80 code variants under expensive structural validation. At a matched evaluation budget, it achieves 7.5\% higher best lDDT than an independent Codex-proposal baseline and also outperforms a random-search controller.
    
    \item \textbf{Trace-based design evidence.} We analyze the resulting intervention traces to identify recurring post-hoc empirical patterns: stable gains are associated with early soft learnable priors and gated refinement, whereas direct geometric perturbations and geometry-conditioned feedback often destabilize training.
\end{itemize}

\section{Related Work}
\subsection{Autonomous AI Research}
LLM-based systems support literature synthesis and hypothesis generation, end-to-end scientific workflows, and biomedical research planning~\cite{lu2024ai,boiko2023autonomous,Swanson2025TheVL}. Execution-grounded benchmarks further evaluate agents on iterative machine-learning experimentation, scientific or data-science code generation, and research-code extensions~\cite{Tian2024SciCodeAR,Huang2023MLAgentBenchEL,Chan2024MLEbenchEM,Huang2024DACodeAD,Edwards2025RExBenchCC}. These studies expose the difficulty of long-horizon implementation and validation, but they do not specialize the loop to protein-model development.

A complementary line couples LLM-generated programs or designs with executable feedback. FunSearch and AlphaEvolve evolve programs, MCTS-AHD applies tree search to heuristic design, and RZ-NAS and ASI-ARCH search model architectures~\cite{RomeraParedes2023MathematicalDF,novikov2025alphaevolve,Zheng2025MonteCT,Ji2025RZNASEL,liu2025alphago}. AgentFold provides a domain-specific instantiation of these general components in a tightly coupled protein-folding codebase, where each proposed variant must be implemented, debugged, trained, and evaluated against multiple structural metrics.

\subsection{Protein Folding}
Protein structure prediction has progressed from MSA-based systems such as AlphaFold2 and RoseTTAFold~\cite{jumper2021highly, Baek2023EfficientAA} to unified complex predictors such as AlphaFold3 and RoseTTAFold All-Atom~\cite{Abramson2024AccurateSP,Krishna2023GeneralizedBM}. Open, trainable platforms including OpenFold and Uni-Fold support method development and reproducible engineering~\cite{ahdritz2024openfold,Li2022UniFoldAO}. Other work explores MSA-free language-model-based prediction with OmegaFold and ESMFold~\cite{Wu2022HighresolutionDN,Lin2022EvolutionaryscalePO}, training efficiency with FastFold and MiniFold~\cite{Cheng2022FastFoldRA,Wohlwend2025MiniFoldSF}, and generative or flow-based formulations such as EigenFold, AlphaFlow/ESMFlow, and SimpleFold~\cite{Jing2023EigenFoldGP,Jing2024AlphaFoldMF,Wang2025SimpleFoldFP}. 

\section{Method}
We view autonomous folding-model development as a \emph{search over code-level interventions} and their measured outcomes. AgentFold is designed to produce two coupled artifacts: (\emph{i}) improved model variants and (\emph{ii}) accumulated design evidence distilled from intervention--outcome traces. We use an MCTS-style tree controller over executable code variants, enabling compute-efficient exploration and controlled comparisons among competing design choices. A self-evolving multi-agent loop proposes and implements edits, evaluates variants, and recovers from failures. Finally, an attribution-and-retrieval stage writes structured intervention artifacts to a database-backed memory, while periodic re-scoring updates node values and refines the search policy. Prompts and templates are provided separately (see Appendix~\ref{app:prompts}).

\subsection{Problem Formulation \& Overview}
Given a base folding model $\mathcal{M}_0$ (ESMFold~\cite{Lin2022EvolutionaryscalePO}), we aim to discover variants $\{\mathcal{M}_t\}_{t=1}^T$ that improve target evaluation metrics and, in parallel, to summarize recurring empirical design patterns $\mathcal{P}=\{P_k\}_{k=1}^{K}$ from repeated intervention evidence. Each iteration logs a structured \emph{intervention trace} that records the parent variant, the typed edit (e.g., priors, refinement control, geometry operations), the code diff, stability signals, and metric deltas, which supports cross-variant attribution and empirical pattern mining.

\begin{figure*}[ht]
    \centering
    \includegraphics[width=\textwidth]{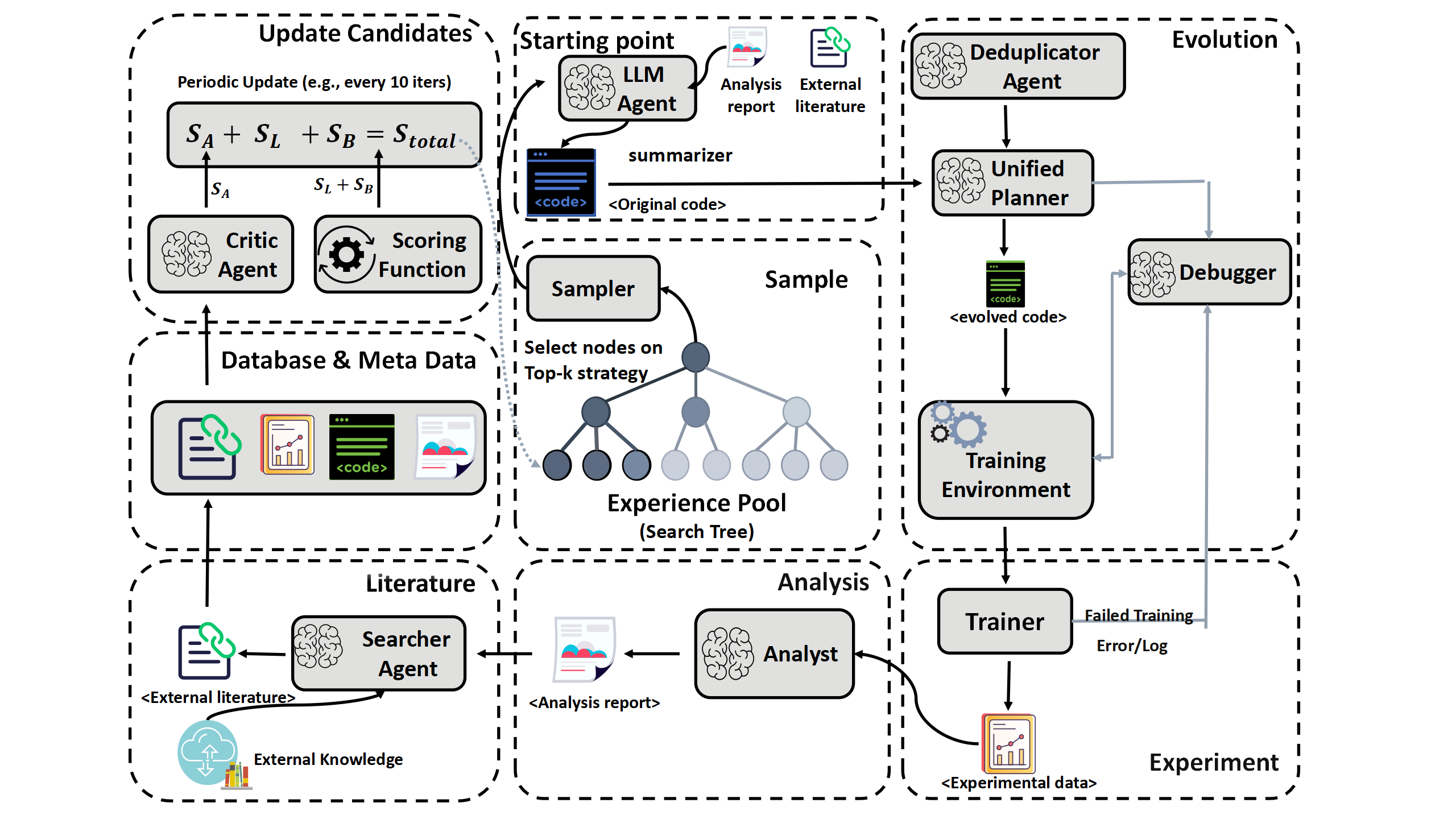}
    \caption{AgentFold system overview. We cast model improvement as MCTS-style search over a code-variant tree, coupling an inner loop (sample $\rightarrow$ evolve $\rightarrow$ run $\rightarrow$ analyze) with a database-backed memory, and an outer periodic update that re-scores candidates to refine the search policy.}
    \label{fig:arch}
\end{figure*}
To address the complexity of the ESMFold codebase, we propose AgentFold, an LLM-based multi-agent framework with an MCTS-style tree controller. As illustrated in Figure~\ref{fig:arch}, AgentFold operates via a dual-loop mechanism:
\begin{itemize}[leftmargin=*]
\item \textbf{Inner Exploration Loop:} A continuous cycle of Sampling, Evolution, Experiment, and Analysis that iteratively generates and verifies new model variants.
\item \textbf{Outer Periodic Update:} A batched update mechanism (e.g., every 10 iterations) that refines the search tree and candidate sets using a composite scoring function.
\end{itemize}

A central Database \& Metadata module serves as an experiment memory: it stores executable code snapshots, code diffs, configurations, logs, and structured attributions, linking them to retrieved literature so that future edits can be proposed and evaluated using accumulated evidence.
\subsection{MCTS-based Dynamic Sampling}
The search process begins with the Experience Pool (Search Tree), which structurally organizes model variants.

\textbf{Top-k Sampling Strategy.} Sampling multiple siblings from the same parent node creates near-controlled comparisons (holding most code constant), supporting attribution of gains or losses to specific intervention types and consolidation of recurring design patterns. At the start of each inner loop, the sampler selects high-scoring nodes together with diverse reference nodes, approximating an exploration--exploitation trade-off.

\textbf{Context Summarization.} The Summarizer prioritizes evidence that is comparable to the current parent node (e.g., similar edit types or failure modes), producing a compact brief that highlights successful outcomes, failed interventions, and empirical patterns currently supported by the accumulated traces.
\subsection{Self-Evolving Agentic Workflow}
The Evolution phase transforms the summarized context into executable code through a specialized agent chain:
\begin{enumerate}[leftmargin=*]
    \item   \textbf{Deduplication.} First, a Deduplicator Agent screens the proposed optimization direction against historical data to prevent redundant experiments.
    \item  \textbf{Unified Planning \& Coding.} Valid proposals are passed to the Unified Planner. Unlike decoupled approaches, this agent is solely responsible for both architectural design and code implementation, reducing interface mismatches between design and implementation.
    \item  \textbf{Interactive Debugging.} The generated code enters the Training Environment. A Debugger agent monitors the process in real time. Upon detecting an error or anomalous log message, the Debugger autonomously interacts with the Unified Planner to iteratively fix syntax or runtime errors until training launches successfully.
\end{enumerate}

\subsection{Attribution, Empirical Pattern Mining \& Knowledge Retrieval}
Once training concludes (or terminates unsuccessfully), the system initiates a two-stage post-processing phase to enrich the \textbf{Database \& Metadata:}

\begin{enumerate}[leftmargin=*]
    \item \textbf{Automated analysis.} The Trainer streams logs to an Analyst agent, which summarizes likely contributors to metric/stability changes and produces a structured report: attribution of deltas to the intervention and an evidence-based update to the current candidate pattern set $\mathcal{P}$ (support, refute and qualify). Reports are persisted to the database.
    \item \textbf{Literature augmentation.} In parallel, a Searcher agent monitors new records, retrieves relevant external literature, and links it to the corresponding interventions and observed failure modes, providing context for subsequent proposals.
\end{enumerate}

\subsection{Periodic Update \& Scoring Mechanism}
Whereas canonical MCTS updates node values after each rollout, AgentFold uses batched periodic updates every 10 iterations because each rollout corresponds to an expensive training/evaluation job. We employ a hybrid evaluation module depicted as the "Update Candidates" block. An algorithmic metric parser and a Critic Agent collaboratively compute the total score $S_{total}(e)$:
\[
S_{\text{total}}(e) = S_L(e) + S_B(e) + S_A(e)
\]
\begin{itemize}[leftmargin=*]
    \item \textbf{Objective metrics ($S_L + S_B$): } The metric parser automatically extracts the loss score ($S_L$) and benchmark score ($S_B$) from the training logs stored in the database.
    \item \textbf{Critic score ($S_A$):} The Critic Agent reviews the intervention rationale and implementation risk (e.g., coherence with prior evidence, clarity of hypothesis, and likelihood of destabilizing training), yielding an agent score $S_A$ used only to prioritize expensive experiments rather than to claim final improvements.
\end{itemize}

\textbf{Tree Refinement.} At the end of each period, these scores are aggregated to update the node values in the Experience Pool. This periodic synchronization allows the global search policy (Top-k strategy) to evolve based on a batched, robust assessment of recent explorations.

\section{Results}
We first describe the benchmark-guided experimental setup, then report overall search behavior and CAMEO2022 development-benchmark performance. We next use targeted metrics to localize where the gains occur, analyze the variant tree to identify recurring empirical design patterns, and finally test the strongest variant through repeated runs, component ablations, and qualitative loop-region cases.

\subsection{Experiment Setup}
\textbf{Baseline and training data.}
We start from a compact ESMFold-derived baseline~\cite{Lin2022EvolutionaryscalePO}, which preserves the sequence, pair, and structure-module interactions needed for controlled folding-model edits while making repeated search feasible (see Appendix~\ref{app:exp_details:model_details}). For training, we sample a 1{,}000-chain mini-dataset from temporally split PDB chains using MMseqs2 cluster-aware weighting and a medium-length preference (see Appendix~\ref{app:exp_details:data_curation}).

\textbf{Evaluation.}
We use CAMEO2022~\cite{Haas2017ContinuousAM} as the development benchmark for scoring variants and allocating search compute. We report backbone lDDT, lDDT, oligomeric GDT-TS, RMSD, and TM-score using OpenStructure~\cite{Biasini2013OpenStructureAI}; NWRS aggregates these metrics relative to a fixed ESMFold baseline for benchmark-guided search ranking (see Appendix~\ref{app:exp_details:metric_def},~\ref{app:exp_details:NWRS}).

\subsection{Search and Overall Performance}
\subsubsection{Quantitative analysis of Monte Carlo tree evolution}
\label{sec:mcts_quant}

\begin{figure}[ht]
    \centering
    \includegraphics[width=1\linewidth]{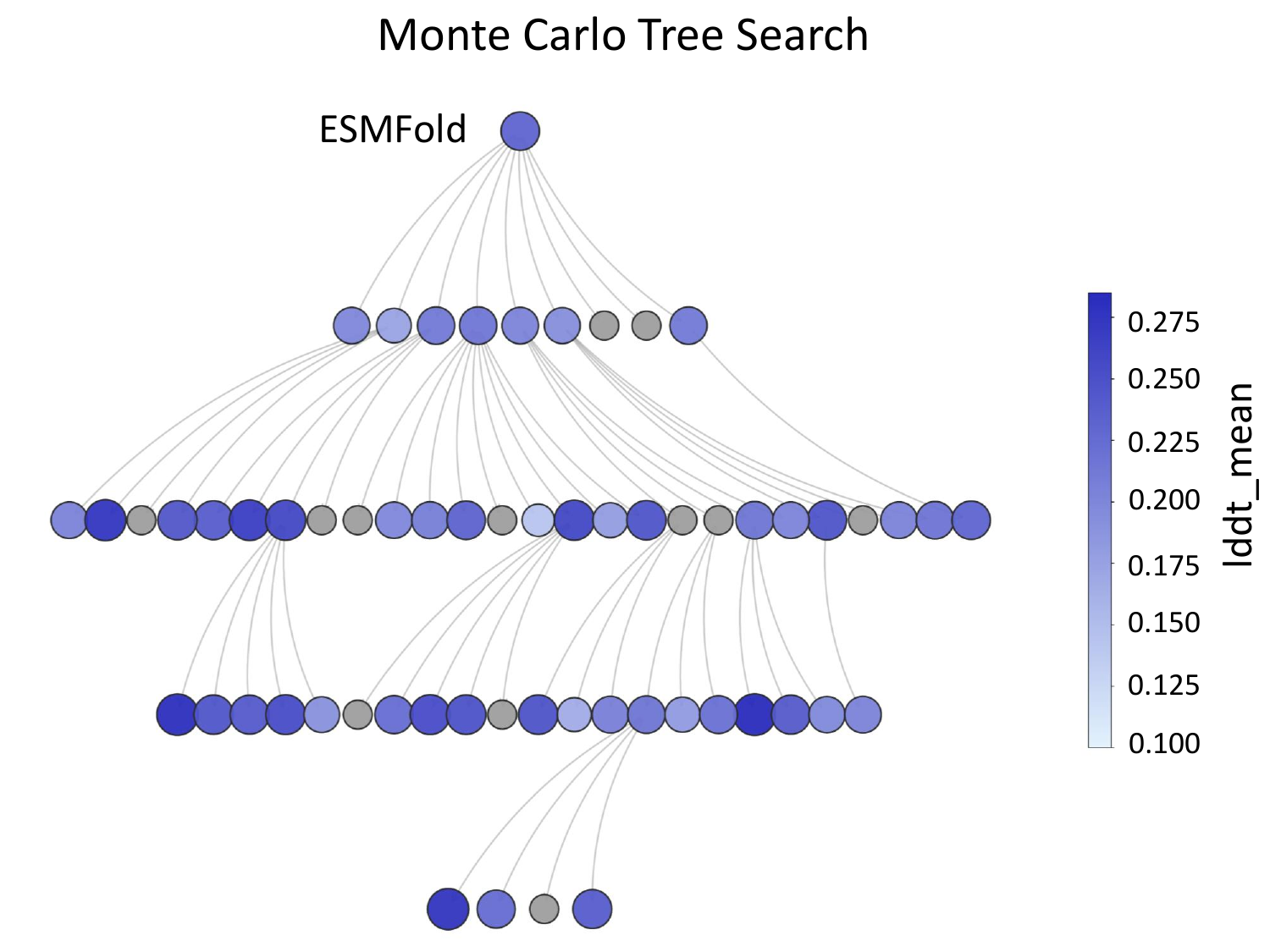}
    \caption{\textbf{MCTS-style tree evolution.} Each node is a sampled variant scored by average lDDT ($\texttt{lddt\_mean}$). Color encodes performance (darker indicates higher $\texttt{lddt\_mean}$); gray marks low-scoring variants with $\texttt{lddt\_mean}<0.1$.}
    \label{fig:MCTREE}
\end{figure}

Figure~\ref{fig:MCTREE} visualizes sampled variants as tree nodes: darker nodes indicate higher mean lDDT, and gray nodes mark low-scoring candidates. The trajectory follows a wide-to-focused pattern. Early iterations sample heterogeneous edits with mixed outcomes, whereas later expansions form denser branches around higher-lDDT variants. The observed trajectory is consistent with the MCTS-style controller reallocating compute toward high-scoring code-variant neighborhoods; we interpret it as descriptive evidence of search behavior, while noting that it does not constitute a controlled comparison against alternative controllers.

\subsubsection{Matched Search-Controller Comparison}
\label{sec:search_baselines}

\begin{table}[t]
\centering
\captionsetup{font=small}
\caption{Matched comparison at 36 evaluations; Top-5 by NWRS.}
\label{tab:search_baselines}
\setlength{\tabcolsep}{3pt}
\scriptsize
\resizebox{\columnwidth}{!}{%
\begin{tabular}{l r r r r r}
\toprule
Method & Best lDDT & Top-5 lDDT & Best NWRS & Top-5 NWRS \\
\midrule
AgentFold & \textbf{0.285} & \textbf{0.267} & \textbf{0.526} & \textbf{0.516} \\
Codex proposals & 0.265 & 0.257 & 0.512 & 0.509 \\
Random controller & 0.260 & 0.242 & 0.510 & 0.506 \\
\bottomrule
\end{tabular}%
}
\end{table}

With 36 evaluations each, AgentFold outperforms two equal-budget baselines. Random control uses the same edit space, models, prompts, checks, training, and evaluator but selects actions randomly. Codex independently generates proposals without the search tree or intervention history; executable candidates use the same pipeline. AgentFold achieves the best and NWRS-selected Top-5 results (Table~\ref{tab:search_baselines}), supporting the integrated search while not isolating individual components.

\subsubsection{Quantitative Results}

\begin{table*}[htbp]
\centering
\captionsetup{font=small}
\caption{CAMEO2022 development-benchmark performance for representative variants. We show the top NWRS variants and variants used in later targeted analyses. The ESMFold row reports absolute mean/median values; other rows report deltas relative to ESMFold. \textbf{Bold} and \underline{underline} mark the largest and second-largest favorable changes among displayed variants.}
\label{tab:variants_mean_median_diff}

\rowcolors{2}{gray!6}{white}
\renewcommand{\arraystretch}{1.15}
\setlength{\tabcolsep}{4pt}
\scriptsize

\resizebox{\linewidth}{!}{%
\begin{tabular}{l c c c c c c}
\toprule
\rowcolor{gray!15}
Variant & NWRS $\uparrow$ & bb\_lddt $\uparrow$ & lddt $\uparrow$ & oligo\_gdtts $\uparrow$ & rmsd $\downarrow$ & tm\_score $\uparrow$ \\
\midrule
esmfold & 0.500 & 0.644/0.651 & 0.232/0.220 & 0.564/0.570 & 7.380/5.358 & 0.648/0.693 \\
esmfold\_struct\_enhanced\_v4 & \textbf{+0.026} & +0.009/+0.010 & \textbf{+0.053}/\textbf{+0.059} & +0.005/-0.003 & +0.082/-0.038 & +0.004/-0.012 \\
esmfold\_struct\_local\_context\_v1 & \underline{+0.020} & +0.002/+0.006 & \underline{+0.049}/+0.044 & -0.001/-0.005 & +0.176/\underline{-0.129} & +0.001/-0.012 \\
esmfold\_struct\_dist\_aware\_v1 & +0.018 & \underline{+0.011}/\textbf{+0.014} & +0.027/+0.024 & \textbf{+0.011}/\textbf{+0.020} & -0.088/\textbf{-0.134} & \textbf{+0.011}/-0.009 \\
esmfold\_struct\_enhanced\_multiscale\_v2 & +0.017 & +0.007/\textbf{+0.014} & +0.045/\underline{+0.049} & +0.004/+0.006 & +0.261/+0.338 & +0.001/-0.019 \\
esmfold\_net\_conformal\_geometric\_attention & +0.017 & -0.006/-0.008 & +0.043/+0.046 & -0.006/\underline{+0.011} & -0.063/+0.240 & -0.005/-0.001 \\
esmfold\_struct\_enhanced\_v1\_dup2 & +0.014 & \textbf{+0.012}/+0.010 & +0.023/+0.025 & \underline{+0.007}/+0.010 & +0.029/-0.038 & \underline{+0.007}/-0.009 \\
esmfold\_struct\_attn\_frame\_v1 & +0.010 & +0.002/+0.007 & +0.016/+0.020 & +0.001/\underline{+0.011} & +0.025/-0.126 & +0.001/-0.013 \\
esmfold\_struct\_enhanced\_v2\_dup3 & +0.007 & +0.006/+0.008 & +0.017/+0.003 & +0.003/+0.004 & \textbf{-0.127}/\underline{-0.129} & +0.003/-0.017 \\
\bottomrule
\end{tabular}%
}
\end{table*}

Table~\ref{tab:variants_mean_median_diff} reports mean/median performance for representative variants, with each non-baseline row shown as a delta relative to ESMFold. Under the CAMEO2022-guided search protocol, all displayed variants improve NWRS (+0.007 to +0.026) and mean lDDT (+0.016 to +0.053), indicating that the search repeatedly finds executable edits with better local structural accuracy rather than a single isolated outlier. The strongest overall variant, \texttt{esmfold\_struct\_enhanced\_v4}, has the largest composite gain (+0.026) and the largest lDDT gain in both mean and median (+0.053/+0.059), while \texttt{esmfold\_struct\_local\_context\_v1} and \path{esmfold_struct_enhanced_multiscale_v2} show similarly local-accuracy-oriented profiles.

The gains are not uniform across global metrics, which is important for interpreting the result. \texttt{esmfold\_struct\_dist\_aware\_v1} gives a smaller lDDT gain than \texttt{esmfold\_struct\_enhanced\_v4} but is more favorable on backbone lDDT, GDT-TS, mean RMSD, and mean TM-score. Conversely, several high-NWRS variants improve lDDT while leaving TM-score nearly unchanged and producing mixed RMSD changes. This pattern shows that AgentFold's improvements are concentrated in local structural accuracy while largely preserving, rather than systematically improving, global fold quality. It also motivates the targeted analyses below, where we separate loop quality, physical plausibility, and contact behavior instead of relying only on a single aggregate score.

\subsubsection{Targeted Evaluation of Inductive Biases}
Each variant encodes a specific inductive bias, but aggregate metrics are insufficient to test whether the intended behavior emerges. We therefore cluster motivations into five recurring goal categories (see Appendix Table~\ref{tab:variant_goals_refined}) and evaluate each goal with targeted metrics. This goal-conditioned analysis supports controlled comparison across variants (reported as $\Delta$ vs.\ ESMFold) and clarifies which biases translate into consistent, measurable gains.

\textbf{Motivation-aspect summary.}
Table~\ref{tab:targeted_eval_summary} merges the targeted loop, physical, and contact evaluations by taking the union of representative variants from these aspects. Each row is annotated by its motivation aspect(s): L denotes loop quality, P denotes physical plausibility, and C denotes contact modeling. The main table keeps two loop metrics, MolProbity for physical plausibility, and two contact metrics in the 12--24 sequence-separation bin; complete targeted metrics are reported separately (see Appendix Tables~\ref{tab:targeted_loop_full}--\ref{tab:targeted_contact_full}).

\begin{table*}[t]
\centering
\captionsetup{font=small}
\caption{Targeted evaluation summary by motivation aspect. The ESMFold row reports absolute means; other rows report changes relative to ESMFold. L/P/C denote loop-quality, physical-plausibility, and contact-modeling motivations. \textbf{Bold} indicates the largest improvement, and \underline{underline} indicates the second largest.}
\label{tab:targeted_eval_summary}
\rowcolors{3}{gray!6}{white}
\renewcommand{\arraystretch}{1.12}
\setlength{\tabcolsep}{3.5pt}
\scriptsize
\resizebox{\linewidth}{!}{%
\begin{tabular}{l c c c c c c}
\toprule
\rowcolor{gray!15}
Variant & Aspect & \multicolumn{2}{c}{Loop} & Physical & \multicolumn{2}{c}{Contact} \\
\cmidrule(lr){3-4}\cmidrule(lr){5-5}\cmidrule(lr){6-7}
\rowcolor{gray!15}
 & & loop lDDT $\uparrow$ & loop bb-lDDT $\uparrow$ & MolProbity $\downarrow$ & Prec$_{12\text{-}24}$ $\uparrow$ & F1$_{12\text{-}24}$ $\uparrow$ \\
\midrule
esmfold & Base & 0.162 & 0.613 & 3.773 & 0.599 & 0.606 \\
esmfold\_struct\_enhanced\_v4 & L/P/C & \underline{+0.060} & \textbf{+0.008} & \textbf{-0.157} & \underline{+0.019} & \underline{+0.010} \\
esmfold\_struct\_local\_context\_v1 & L & \textbf{+0.063} & +0.002 & -- & -- & -- \\
esmfold\_struct\_enhanced\_v1\_dup2 & L/C & +0.031 & \underline{+0.007} & -- & \textbf{+0.020} & \textbf{+0.013} \\
esmfold\_struct\_attn\_frame\_v1 & L/P & +0.025 & +0.001 & \underline{-0.049} & -- & -- \\
esmfold\_struct\_enhanced\_multiscale\_v2 & L/P/C & +0.056 & +0.002 & -0.043 & +0.009 & +0.007 \\
esmfold\_struct\_enhanced\_v2\_dup3 & L/C & +0.023 & +0.002 & -- & +0.013 & +0.006 \\
\bottomrule
\end{tabular}%
}
\end{table*}

Table~\ref{tab:targeted_eval_summary} decomposes the aggregate gains in Table~\ref{tab:variants_mean_median_diff}. The loop columns show that loop-oriented improvements are concentrated in loop lDDT: \texttt{esmfold\_struct\_local\_context\_v1} has the largest loop-lDDT gain (+0.063), whereas \texttt{esmfold\_struct\_enhanced\_v4} has the largest loop backbone-lDDT gain (+0.008). For physical plausibility, \texttt{esmfold\_struct\_enhanced\_v4} achieves the largest MolProbity reduction (-0.157), with \texttt{esmfold\_struct\_attn\_frame\_v1} showing a smaller reduction (-0.049). Contact gains are more selective: in the 12--24 separation bin, \texttt{esmfold\_struct\_enhanced\_v1\_dup2} yields the largest precision and F1 gains (+0.020/+0.013), while \texttt{esmfold\_struct\_enhanced\_v4} yields comparable gains (+0.019/+0.010). Together, the targeted metrics support the same conclusion as Table~\ref{tab:variants_mean_median_diff}: AgentFold's largest gains are local and medium-range rather than broad global-fold improvements. See Appendix Tables~\ref{tab:targeted_loop_full}--\ref{tab:targeted_contact_full} for the complete targeted metrics.

\subsection{Analysis}
We analyze the variant tree to assess whether the gains reflect recurring empirical design patterns rather than capacity effects. This analysis is descriptive: it compares successful and failed edits in the same search tree and summarizes patterns that repeatedly co-occur with stable or unstable outcomes.

\subsubsection{Evolutionary Analysis}
\label{sec:evolutionary_analysis}
\begin{figure}[t]
\centering
\captionsetup{font=small}
\includegraphics[width=\linewidth]{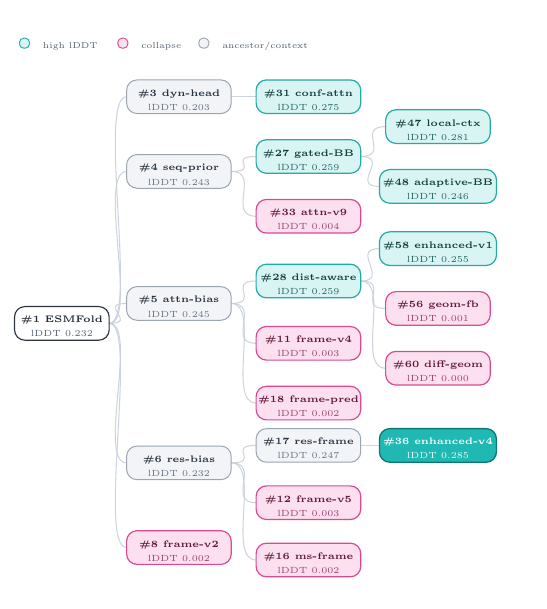}
\caption{Selected variant subtree used in the evolutionary analysis. Colors distinguish high-lDDT variants, collapse cases, and ancestor/context nodes; each node reports mean lDDT.}
\label{fig:evolution_tree_maintext}
\end{figure}

\paragraph{Variant-tree trends by mean lDDT.}
Figure~\ref{fig:evolution_tree_maintext} summarizes the selected subtree used for this analysis, with each node annotated by mean lDDT. The high-performing region is not defined by a single module name; instead, strong variants such as \#36 (\texttt{esmfold\_struct\_enhanced\_v4}) and \#47 (\texttt{esmfold\_struct\_local\_context\_v1}) share a similar placement strategy: they add \emph{soft, learnable priors} before coordinates are instantiated. In contrast, severe failures such as \#60 (\texttt{esmfold\_net\_differential\_geometry}) rely on more direct geometric perturbations after structural information is already being formed. The resulting pattern set $\mathcal{P}$ contains three post-hoc empirical categories rather than theoretical laws: \emph{(P1) Bias before geometry}, \emph{(P2) Multiplicative refinement}, and \emph{(P3) Avoid geometry-to-attention feedback}. The corresponding agent-report evidence is summarized separately (see Appendix Table~\ref{tab:pattern_report_evidence}). P1 is plausible because early pair/IPA biases steer attention before coordinates enter the recycling loop, while late frame-level offsets perturb an already coupled rigid-update process. P2 is less intrusive than additive forcing because gates scale update magnitudes and can damp uncertain regions rather than imposing a fixed geometric displacement. P3 reflects a failure mode in which geometry-derived signals are fed back into attention or frame updates; when initial geometry is inaccurate, this can amplify the error across subsequent refinement steps. The highest-NWRS composite design \#36 combines smooth IPA biasing, gated updates, and chunk-boundary attention while leaving the core IPA$\rightarrow$frames$\rightarrow$FAPE loop intact, which may explain why it improves local metrics without disrupting global fold quality.

The tree suggests an empirical design heuristic: stable improvements are associated with \emph{early, learnable priors} and \emph{multiplicative control} of refinement, whereas \emph{direct geometric forcing} and \emph{geometry-conditioned feedback} are associated with collapse in evaluation. This is consistent with the quantitative results above: successful edits tend to steer attention or update magnitudes, while failed edits more often impose geometry directly.

\subsubsection{Parameter Analysis}
\paragraph{Parameter-efficiency of gains.}
High-NWRS variants remain close to the $22.61$M-parameter ESMFold baseline. The highest-NWRS model \#36 has $22.856$M parameters, an increase of only $\sim 1.1\%$, and several strong variants add less than $0.1\%$. For example, \#47 adds approximately $0.015$M parameters yet reaches the second-highest NWRS in Table~\ref{tab:variants_mean_median_diff}, and \#28 slightly reduces the parameter count while improving several backbone/global metrics. Conversely, larger variants are not reliably better: \#24 and \#40 have $28.46$M parameters and \#57 has $32.49$M, but they do not dominate the compact high-NWRS variants; \#60 collapses despite having $31.03$M parameters. These comparisons suggest that the gains are better explained by the placement of biases and gates than by raw capacity.

\subsection{Ablation Study}
\begin{figure}[t]
\centering
\captionsetup{font=small}
\includegraphics[width=\linewidth]{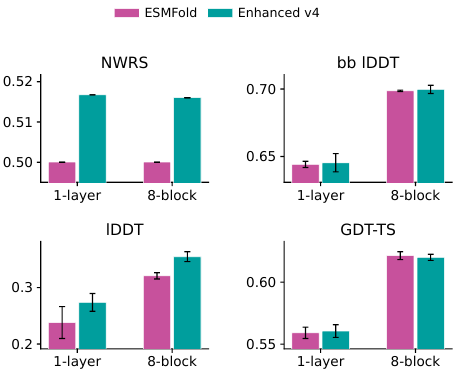}
\caption{Robustness under repeated runs and deeper Folding Trunks. Matched repeated-run settings are used; RMSD is omitted due to its different scale. Error bars denote standard deviations.}
\label{fig:ablation_robustness}
\end{figure}

Figure~\ref{fig:ablation_robustness} shows that \texttt{esmfold\_struct\_enhanced\_v4} preserves its lDDT advantage under repeated runs and an 8-block Folding Trunk. In the 1-layer setting, mean lDDT increases from $0.238$ to $0.274$; with 8 trunk blocks, it increases from $0.321$ to $0.355$.  These follow-up runs use matched repeated-run baselines; therefore, their ESMFold means are not expected to exactly match the single-run Table~\ref{tab:variants_mean_median_diff} baseline.

\begin{figure}[t]
\centering
\captionsetup{font=small}
\includegraphics[width=\linewidth]{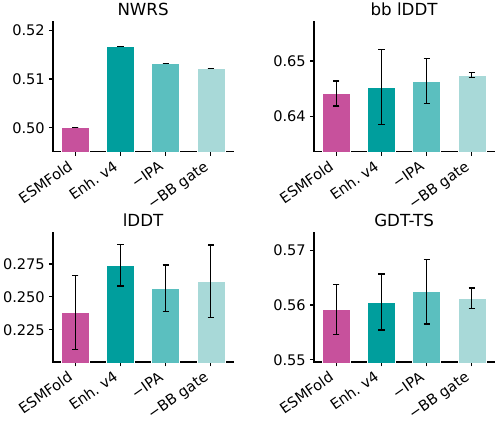}
\caption{Component ablation of \texttt{esmfold\_struct\_enhanced\_v4}. Each panel shows one metric; RMSD is omitted due to its different scale. Error bars denote standard deviations.}
\label{fig:esmfold_struct_enhanced_v4_ablation}
\end{figure}

Figure~\ref{fig:esmfold_struct_enhanced_v4_ablation} tests whether the highest-NWRS variant is driven by a single component. Removing the IPA bias or BackboneUpdate gating lowers mean lDDT by $0.017$ and $0.012$, respectively. Both ablated variants remain competitive with ESMFold on some metrics, but neither recovers the full lDDT gain, indicating that the bias and gating mechanisms are complementary rather than interchangeable. Mechanistically, the IPA bias changes where information is routed during attention, whereas BackboneUpdate gating controls how strongly the resulting update is applied; removing either weakens a different part of the refinement pathway.

\begin{figure}[ht]
    \centering
    \includegraphics[width=1\linewidth]{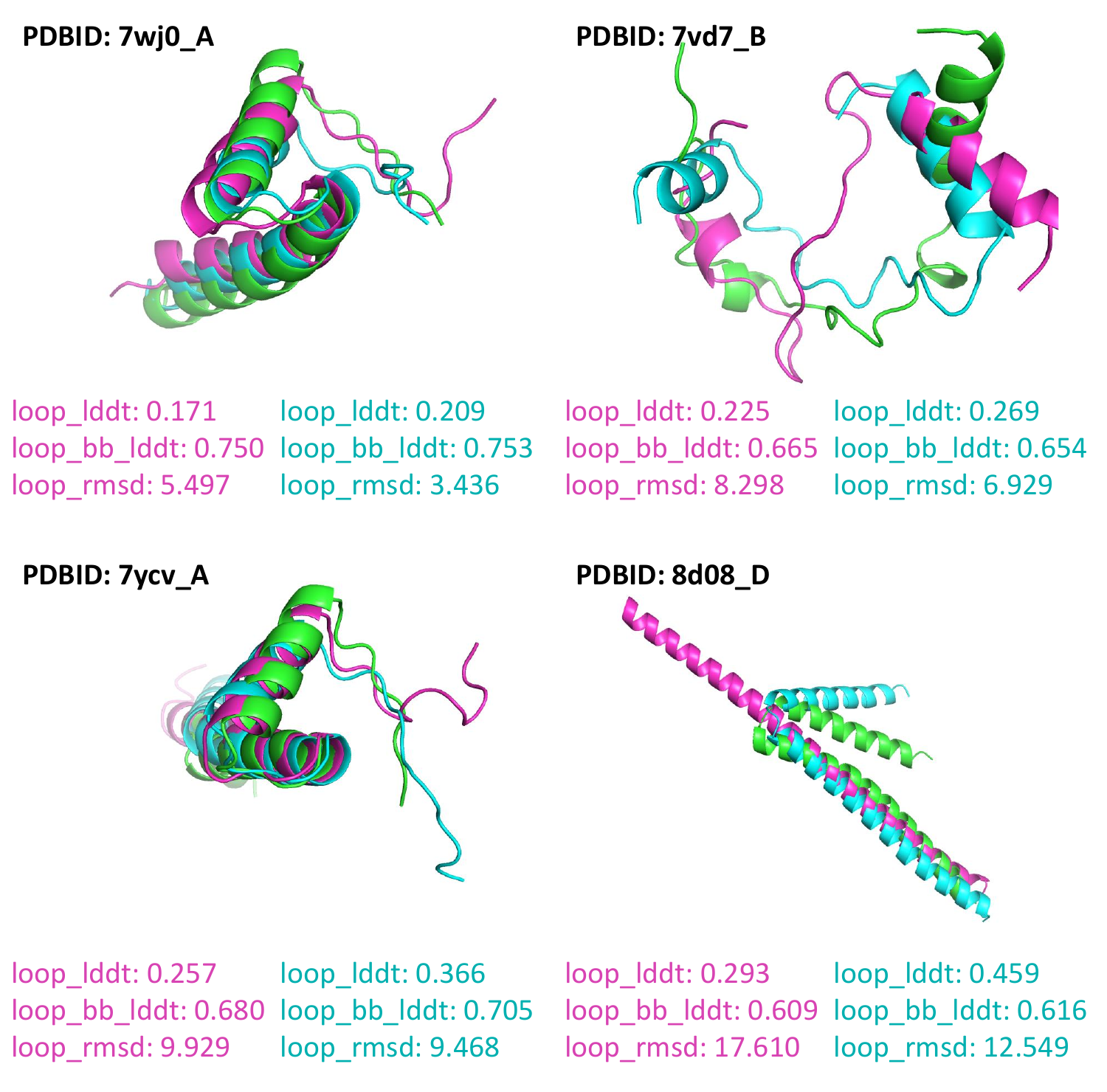}
    \caption{\textbf{Loop-region case studies on four CAMEO targets.}
    Superpositions of ground truth (green), ESMFold (magenta), and \texttt{esmfold\_struct\_enhanced\_v4} (cyan) are shown for 7wj0\_A, 7vd7\_B, 7ycv\_A, and 8d08\_D.
    Text in each panel reports loop lDDT, loop backbone lDDT, and loop RMSD for ESMFold and \texttt{esmfold\_struct\_enhanced\_v4}.
    \texttt{esmfold\_struct\_enhanced\_v4} improves loop placement and backbone alignment in most cases, while 7vd7\_B illustrates a residual metric trade-off.}
    \label{fig:case_loop}
\end{figure}

\subsection{Case Study}
We use \texttt{esmfold\_struct\_enhanced\_v4} as a representative case because it achieves the largest local-accuracy gains while remaining close to the baseline architecture; implementation details are summarized separately (see Appendix~\ref{app:Architecture}).

\paragraph{Loop-region improvement.}
Loops are challenging due to weak constraints and high flexibility.
Figure~\ref{fig:case_loop} visualizes four representative loop-region cases (PDB IDs: 7wj0\_A, 7vd7\_B, 7ycv\_A, and 8d08\_D).
Compared with ESMFold, \texttt{esmfold\_struct\_enhanced\_v4}
reduces loop RMSD in all four cases and consistently improves
loop lDDT, while loop backbone lDDT increases in three of the
four targets. 

\paragraph{Mechanistic interpretation.}
The architecture comparison (see Appendix Figure~\ref{fig:esmfold_variant}) shows that the variant inserts IPA-side biasing while preserving the downstream geometric heads.
Together with the ablation results, the qualitative examples are consistent with the quantitative trend: IPA biasing and BackboneUpdate gating appear to improve flexible-loop placement without systematically changing global topology.

\section{Conclusion}
We present AgentFold, a multi-agent framework that formulates folding-model development as closed-loop search over executable code variants. Starting from ESMFold, AgentFold identifies parameter-efficient variants with consistent gains, primarily in local structural accuracy, while largely preserving global fold quality. The intervention traces further suggest recurring empirical design patterns: early soft learnable priors and gated refinement are associated with more stable gains in our search, whereas direct geometric perturbations and geometry-conditioned feedback often destabilize training.

\section*{Limitations}
Our evidence is limited to a one-block, compact ESMFold-derived codebase, a 1{,}000-chain training subset, and CAMEO2022 development-benchmark evaluation; transfer to stronger folding systems and broader biological settings remains unverified.

\paragraph{Future work.} Extending the discovered interventions to larger and multi-chain systems requires model-specific edit interfaces, chain-aware representations, interface-sensitive objectives, retraining, and evaluation. Cross-domain use similarly requires a domain-specific codebase, evaluator, reward, and failure-analysis loop. We leave these extensions to future work.

\section*{Acknowledgments}

The authors thank Zehong Wang (University of Notre Dame) for helpful discussions and suggestions. This work was supported by the National Natural Science Foundation of China (Grant Nos.~62425204, U22A2037, 62450002, and 62432011). Jiangyu Chen and Tianfan Fu were supported by the Young Scientists Fund (C Class) of the National Natural Science Foundation of China (Grant No.~62506154), the Fundamental Research Funds for the Central Universities, the Nanjing University International Collaboration Initiative (Grant No.~020214380129), and the ``111 Center'' (No.~B26023).

\section*{Ethical Considerations}
AgentFold aims to improve protein folding models through closed-loop code search. While better structure prediction can support biological and medical research, increased AI-for-biology capability may also introduce dual-use risks. Responsible release, careful evaluation, and human oversight are therefore important.

% In the unusual situation where you want a paper to appear in the
% references without citing it in the main text, use \nocite
% \nocite{langley00}

% \bibliographystyle{acl_natbib}
\bibliography{AgentFold_example_paper_updated}

%%%%%%%%%%%%%%%%%%%%%%%%%%%%%%%%%%%%%%%%%%%%%%%%%%%%%%%%%%%%%%%%%%%%%%%%%%%%%%%
%%%%%%%%%%%%%%%%%%%%%%%%%%%%%%%%%%%%%%%%%%%%%%%%%%%%%%%%%%%%%%%%%%%%%%%%%%%%%%%
% APPENDIX
%%%%%%%%%%%%%%%%%%%%%%%%%%%%%%%%%%%%%%%%%%%%%%%%%%%%%%%%%%%%%%%%%%%%%%%%%%%%%%%
%%%%%%%%%%%%%%%%%%%%%%%%%%%%%%%%%%%%%%%%%%%%%%%%%%%%%%%%%%%%%%%%%%%%%%%%%%%%%%%
\newpage
\appendix
\onecolumn
\section{Experiment Details}\label{app:exp_details}
\subsection{Model details}\label{app:exp_details:model_details}

\paragraph{One-layer Folding Trunk for large-scale exploration.}
To support large-scale architectural search under a fixed compute budget, we instantiate ESMFold's Folding Trunk (Evoformer-style trunk) with a single trunk block in all experiments unless noted otherwise. This reduces the per-variant training/evaluation cost and enables substantially broader exploration. Crucially, we only modify trunk \emph{depth}: all trunk operators (e.g., triangular multiplicative updates and triangle attention) are unchanged.

\paragraph{Codebase refactoring (packaging only; no behavioral change).}
For reproducibility and ease of auditing, we refactored the ESMFold codebase by consolidating core components that were previously spread across multiple files into a single implementation file. The consolidated module includes (i) the \emph{Structure Module} (IPA, backbone updates, and torsion/frame utilities) and (ii) the trunk components used in our experiments (triangle multiplicative updates, triangle attention, and sequence--pair communication layers). This is a packaging-only change: the architecture, parameterization, and numerical behavior remain identical to the original implementation.

\paragraph{Training setup.}
Unless otherwise noted, variants are trained for 150 epochs with Adam, batch size 8, and a peak learning rate of \(1\times 10^{-3}\). The learning-rate schedule uses a warmup start value of \(0\), linear warmup for 1{,}000 steps, delayed decay after 50{,}000 steps, and multiplicative decay by a factor of \(0.95\) every 50{,}000 steps thereafter. We keep these training hyperparameters fixed across variants so that performance differences primarily reflect architectural interventions rather than per-variant hyperparameter tuning.

\subsection{Mini-data curation}\label{app:exp_details:data_curation}
Our training data are derived from the Protein Data Bank (PDB) at the level of single protein chains. To reduce redundancy, we cluster chains by sequence identity using a minimum identity threshold of \(\,0.4\,\) with MMseqs2~\cite{Steinegger2017MMseqs2ES}, and treat each cluster as a sequence family of size \(\lvert\mathcal{C}\rvert\). We then construct a fixed-size subset of \(1{,}000\) chains via weighted stochastic sampling, where each chain is sampled with probability proportional to an inverse family-size term \(\,1/\lvert\mathcal{C}\rvert\,\) (to down-weight over-represented families) and a length-dependent factor that favors moderate-length sequences,
\[
p_i \propto \frac{1}{\lvert\mathcal{C}_i\rvert}\cdot \frac{1}{512}\,\mathrm{clip}(L_i,256,512).
\]
This procedure yields a more diverse training set while controlling both redundancy and sequence-length distribution. 

\subsection{Artifact licenses and terms}\label{app:artifact_licenses}
We use publicly available research artifacts under their respective licenses and terms of use, including the ESMFold/ESM model code and weights, OpenStructure, MMseqs2, PDB-derived structures, and CAMEO2022 evaluation data. We cite the original creators of these artifacts in the relevant method and experiment sections. Our use of these artifacts is limited to research on protein-structure modeling and evaluation, consistent with their intended research use. We do not redistribute restricted benchmark or structure data in this paper; any released code or model variants should be distributed under terms compatible with the corresponding upstream artifacts.

\subsection{Metric definitions}\label{app:exp_details:metric_def}
We report standard structure-evaluation metrics as implemented in OpenStructure~\cite{Biasini2013OpenStructureAI}. Below we summarize the definitions used throughout the paper. Let the target (native) structure be denoted by \(\mathbf{r}^{\text{target}}\) and the predicted model by \(\mathbf{r}^{\text{model}}\).

\paragraph{lDDT (Local Distance Difference Test).}
lDDT is a superposition-free local accuracy metric that evaluates agreement of inter-atomic distances within a local neighborhood. Given a set of considered atom pairs \(\{(a,b)\}\) (typically restricted to pairs within a neighborhood radius, e.g., 15~\AA\ in the target), define \(d_i^{\text{model}}\) and \(d_i^{\text{target}}\) as the distances of the \(i\)-th considered pair in the model and target, respectively. With threshold set \(\mathcal{T}=\{0.5,1.0,2.0,4.0\}\) (in \AA), we compute
\begin{equation}
\mathrm{lDDT}
=
\frac{1}{N}\sum_{i=1}^{N}\frac{1}{|\mathcal{T}|}\sum_{\tau\in\mathcal{T}}
\mathbb{1}\!\left[\left|d_i^{\text{model}}-d_i^{\text{target}}\right|<\tau\right],
\end{equation}
where \(N\) is the number of considered atom-pair distances and \(\mathbb{1}[\cdot]\) is the indicator function. Higher is better.

\paragraph{Backbone lDDT (\texttt{bb\_lddt}).}
Backbone lDDT is the lDDT score computed using only backbone atoms (e.g., \(N\), \(C_\alpha\), \(C\), \(O\); or \(C_\alpha\)-only depending on the evaluation setting):
\begin{equation}
\texttt{bb\_lddt}=\mathrm{lDDT}_{\text{backbone only}}.
\end{equation}

\paragraph{GDT-TS (Global Distance Test--Total Score).}
GDT-TS is a superposition-based global similarity metric defined as the mean of GDT scores at multiple distance cutoffs:
\begin{equation}
\mathrm{GDT\_TS}
=
\frac{1}{4}\left(\mathrm{GDT}_{1\text{\AA}}+\mathrm{GDT}_{2\text{\AA}}+\mathrm{GDT}_{4\text{\AA}}+\mathrm{GDT}_{8\text{\AA}}\right),
\end{equation}
where, for a cutoff \(d\), the corresponding term is
\begin{equation}
\mathrm{GDT}_{d}
=
\frac{1}{L}\sum_{i=1}^{L}\mathbb{1}\!\left[\left\lVert\mathbf{r}^{\text{model}}_{i}-\mathbf{r}^{\text{target}}_{i}\right\rVert_2<d\right].
\label{eq:gdt-cutoff}
\end{equation}
Here \(L\) is the number of aligned residues (typically using \(C_\alpha\) atoms) and the comparison is performed after an optimal rigid-body superposition.

\paragraph{Oligomeric GDT-TS (\texttt{oligo\_gdtts}).}
For oligomeric targets, we analogously compute GDT-TS on the multi-chain complex after an optimal superposition that accounts for all chains:
\begin{equation}
\texttt{oligo\_gdtts}
=
\frac{1}{4}\left(\mathrm{oligo\_GDT}_{1\text{\AA}}+\mathrm{oligo\_GDT}_{2\text{\AA}}+\mathrm{oligo\_GDT}_{4\text{\AA}}+\mathrm{oligo\_GDT}_{8\text{\AA}}\right),
\end{equation}
where each \(\mathrm{oligo\_GDT}_{d}\) is computed as in Eq.~\ref{eq:gdt-cutoff} but on the oligomeric complex under the corresponding evaluation protocol.

\paragraph{RMSD (Root-Mean-Square Deviation).}
RMSD measures the average Euclidean deviation between corresponding atoms after optimal rigid-body alignment:
\begin{equation}
\mathrm{RMSD}
=
\sqrt{\frac{1}{N}\sum_{i=1}^{N}\left\lVert\mathbf{r}^{\text{model}}_{i}-\mathbf{r}^{\text{target}}_{i}\right\rVert_2^{2}},
\end{equation}
where \(N\) is the number of matched atoms used for the superposition. Lower is better.

\paragraph{TM-score (Template Modeling score).}
TM-score is a length-normalized global similarity metric computed after alignment:
\begin{equation}
\mathrm{TM\mbox{-}score}
=
\max\left\{
\frac{1}{L_{\text{target}}}
\sum_{i=1}^{L_{\text{aligned}}}
\frac{1}{1+\left(d_i/d_0\right)^2}
\right\},
\end{equation}
where \(L_{\text{target}}\) is the target length, \(L_{\text{aligned}}\) is the number of aligned residues, \(d_i\) is the distance between the \(i\)-th aligned \(C_\alpha\) pair after superposition, and \(d_0\) is a length-dependent normalization constant:
\begin{equation}
d_0 = 1.24\sqrt[3]{L_{\text{target}}-15}-1.8.
\end{equation}
Higher is better.

\paragraph{Targeted loop, contact, and physical metrics.}
For loop-region evaluation, we restrict the residue or atom set to predicted loop regions and compute loop lDDT, loop backbone lDDT, and loop RMSD using the corresponding definitions above on that subset. For contact evaluation, residue pairs are grouped by sequence separation bins (\(0\text{--}6\), \(6\text{--}12\), \(12\text{--}24\), and \(\geq 24\)); precision is \(\mathrm{TP}/(\mathrm{TP}+\mathrm{FP})\), recall is \(\mathrm{TP}/(\mathrm{TP}+\mathrm{FN})\), and F1 is \(2PR/(P+R)\). For physical plausibility, MolProbity score, clashscore, Ramachandran outlier rate, rotamer outlier rate, \(C_\beta\) deviations, RMS bond-length deviations, and RMS angle deviations are reported by the structural validation pipeline. Lower is better for these physical-error metrics, while Ramachandran favored residues are reported as a higher-is-better percentage.
\subsection{Normalized Weighted Relative Score (NWRS)} \label{app:exp_details:NWRS}
\label{app:nwrs}

To summarize overall performance with a single scalar, we define the \emph{Normalized Weighted Relative Score} (NWRS). This metric is a weighted, baseline-normalized aggregate over multiple evaluation metrics. NWRS maps a predefined baseline model to a score of $0.5$ and scales other models proportionally, capped at a maximum of $1.0$.

\paragraph{Inputs.}
For a given model, we compute the mean and median across the evaluation set for the following metrics: $\texttt{bb\_lddt}$, $\texttt{lddt}$, $\texttt{oligo\_gdtts}$, $\texttt{rmsd}$, and $\texttt{tm\_score}$. Let $m \in \mathcal{M}$ index the set of ten aggregated metrics:
\begin{equation}
\begin{aligned}
    \mathcal{M} = \big\{
    &\texttt{bb\_lddt\_mean}, \texttt{bb\_lddt\_median}, \\
    &\texttt{lddt\_mean}, \texttt{lddt\_median}, \\
    &\texttt{oligo\_gdtts\_mean}, \texttt{oligo\_gdtts\_median}, \\
    &\texttt{rmsd\_mean}, \texttt{rmsd\_median}, \\
    &\texttt{tm\_score\_mean}, \texttt{tm\_score\_median}
    \big\}.
\end{aligned}
\end{equation}
We denote the model's value for metric $m$ by $x_m$ and the baseline value by $b_m$.

\paragraph{Metric Directions.}
We unify all metrics such that a larger value indicates better performance. We define a direction indicator $s_m \in \{+1, -1\}$, where $s_m = +1$ denotes a \emph{positive} metric (higher is better) and $s_m = -1$ denotes a \emph{negative} metric (lower is better). Specifically:
\begin{equation}
    s_m =
    \begin{cases}
        +1, & \text{if } m \in \mathcal{M} \setminus \{\texttt{rmsd\_mean}, \texttt{rmsd\_median}\}, \\
        -1, & \text{if } m \in \{\texttt{rmsd\_mean}, \texttt{rmsd\_median}\}.
    \end{cases}
\end{equation}
Here, all metrics except RMSD are treated as positive.

\paragraph{Relative Performance Transform.}
We convert each raw metric value into a baseline-relative score $r_m$, where $r_m > 1$ indicates an improvement over the baseline:
\begin{equation}
    r_m =
    \begin{cases}
        x_m / b_m, & \text{if } s_m = +1, \\
        b_m / x_m, & \text{if } s_m = -1.
    \end{cases}
    \label{eq:nwrs-relative}
\end{equation}

\paragraph{Weighted Aggregation and Scaling.}
Given nonnegative weights $\{w_m\}_{m \in \mathcal{M}}$ such that $\sum_{m \in \mathcal{M}} w_m = 1$, the composite score is defined as:
\begin{equation}
    \mathrm{NWRS} = \min\left(1, \; \frac{1}{2} \sum_{m \in \mathcal{M}} w_m \, r_m \right).
    \label{eq:nwrs}
\end{equation}
By construction, if a model matches the baseline exactly ($x_m = b_m$ for all $m$), then $r_m = 1$ and $\mathrm{NWRS} = 0.5$.

\paragraph{Weights and Baseline Values.}
We employ uniform weights across the ten metrics, setting $w_m = 0.1$ for all $m \in \mathcal{M}$. The fixed baseline vector $\{b_m\}$ is defined as follows:
\begin{equation}
\begin{aligned}
    &\texttt{bb\_lddt}       &&: \text{mean} = 0.644, \quad \text{median} = 0.651, \\
    &\texttt{lddt}           &&: \text{mean} = 0.232, \quad \text{median} = 0.220, \\
    &\texttt{oligo\_gdtts}   &&: \text{mean} = 0.564, \quad \text{median} = 0.570, \\
    &\texttt{rmsd}           &&: \text{mean} = 7.380,  \quad \text{median} = 5.358, \\
    &\texttt{tm\_score}      &&: \text{mean} = 0.648, \quad \text{median} = 0.693.
\end{aligned}
\end{equation}
For the ablation study, NWRS is recomputed with the setting-matched ESMFold baseline rather than this fixed main-ranking baseline. This matched-baseline variant preserves the same formula and weights, but maps ESMFold to $0.500$ within each ablation setting.
For numerical stability in Eq.~\eqref{eq:nwrs-relative}, we require $b_m \neq 0$ for all $m$, and $x_m > 0$ for negative metrics (RMSD) to avoid division by zero.
%%%%%%%%%%%%%%%%%%%%%%%%%%%%%%%%%%%%%%%%%%%%%%%%%%%%%%%%%%%%%%%%%%%%%%%%%%%%%%%
%%%%%%%%%%%%%%%%%%%%%%%%%%%%%%%%%%%%%%%%%%%%%%%%%%%%%%%%%%%%%%%%%%%%%%%%%%%%%%%
\section{Results Details} \label{app:re_details}
\subsection{Motivation taxonomy } \label{app:motivation}
\begin{table}[h]
\centering
\caption{Motivation taxonomy of proposed variants. We group each variant's stated goal into five high-level categories (global improvement, loop quality, physical plausibility, long-range contact, and long-sequence quality), and further refine each category by its specific objective. Categories are not mutually exclusive; counts indicate how many variant motivations fall into each objective.} 
\label{tab:variant_goals_refined}
\begin{tabular}{llr} 
\toprule
\textbf{Main Goal} & \textbf{Specific Objective} & \textbf{Count} \\
\midrule
\multirow{2}{*}{\textbf{Global Improvement}} 
& pLDDT/LDDT metric & 41 \\
& RMSD reduction & 21 \\
\midrule
\textbf{Loop Quality} 
& Loop region prediction & 42 \\
\midrule
\multirow{2}{*}{\textbf{Physical Plausibility}} 
& Regularization constraints & 31 \\
& Torsion angle constraints & 22 \\
\midrule
\textbf{Long-range Contact} 
& Long-range contact modeling & 28 \\
\midrule
\textbf{Long Sequence Quality} 
& Long-Sequence TM score & 26 \\
\bottomrule
\end{tabular}
\end{table}

\subsection{Agent-Report Evidence for Empirical Patterns}
\label{app:pattern_report_evidence}

\begin{table*}[t]
\centering
\captionsetup{font=small}
\caption{Representative agent-report evidence used to derive the empirical pattern set $\mathcal{P}$ in Section~\ref{sec:evolutionary_analysis}. The evidence is summarized from the stored intervention reports in \texttt{my\_tree\_data\_dup.json}; it is descriptive rather than causal proof.}
\label{tab:pattern_report_evidence}
\renewcommand{\arraystretch}{1.15}
\setlength{\tabcolsep}{4pt}
\scriptsize
\resizebox{\textwidth}{!}{%
\begin{tabular}{p{0.13\textwidth} p{0.22\textwidth} p{0.43\textwidth} p{0.18\textwidth}}
\toprule
\textbf{Pattern} & \textbf{Representative variants} & \textbf{Evidence from stored reports} & \textbf{Outcome signal} \\
\midrule
P1: Bias before geometry &
\#28 \texttt{dist\_aware\_v1}; \#47 \texttt{local\_context\_v1} &
Reports describe these variants as injecting information into the initial pair representation or IPA logits before coordinates are produced. The reports attribute gains to making residue-pair relations more learnable without directly changing rigid-frame updates. &
\#28 improves backbone/global metrics; \#47 achieves the largest loop-lDDT gain in Table~\ref{tab:targeted_eval_summary}. \\
\midrule
P1 failure contrast &
\#11 \texttt{frame\_reg\_v4}; \#18 \texttt{enhanced\_frame\_pred\_v1} &
Reports note that fixed frame/torsion biases are added directly to BackboneUpdate or combined with attention biases after structural refinement is already coupled. The stored analyses describe collapse or destructive interaction despite plausible motivations. &
Both variants have near-zero lDDT and high RMSD in the search logs, indicating failed folding behavior. \\
\midrule
P2: Multiplicative refinement &
\#36 \texttt{enhanced\_v4}; \#48 \texttt{adaptive\_backbone\_v1} &
Reports describe sigmoid or confidence-related scaling of backbone updates. These mechanisms modulate update magnitude rather than adding a fixed displacement, allowing uncertain regions to be dampened. &
\#36 is the highest-NWRS variant; \#48 improves backbone lDDT and RMSD relative to ESMFold in the stored report. \\
\midrule
P3: Avoid geometry-to-attention feedback &
\#18 \texttt{enhanced\_frame\_pred}; \#57 \texttt{geom\_alg\_phys} &
Reports describe variants that feed structure/geometric features into attention or replace IPA with geometry-heavy attention modules. The reports emphasize that such feedback can introduce conflicting optimization signals when geometric features are immature or noisy. &
\#18 collapses; \#57 is much larger but does not dominate compact variants and shows weaker all-atom/local performance than \#36. \\
\midrule
Hard geometric perturbation failure &
\#60 \texttt{differential\_geometry} &
The report states that curvature/torsion features are computed from sequence features and then fed back into the standard IPA path, while the intended geometric mechanism is not realized. The stored evaluation records a catastrophic failure. &
bb-lDDT $=0.015$, lDDT $=0.000$, TM-score $=0.091$ in the stored test record. \\
\bottomrule
\end{tabular}%
}
\end{table*}

\FloatBarrier
\subsection{Targeted Evaluation Details}
\label{app:targeted_eval_details}

\begin{table}[!htbp]
\centering
\captionsetup{font=small}
\caption{Complete loop-region metrics for the motivation-aspect summary. The ESMFold row reports absolute means; other rows report changes relative to ESMFold.}
\label{tab:targeted_loop_full}
\rowcolors{3}{gray!6}{white}
\renewcommand{\arraystretch}{1.12}
\setlength{\tabcolsep}{4pt}
\scriptsize
\resizebox{\linewidth}{!}{%
\begin{tabular}{l c c c c}
\toprule
\rowcolor{gray!15}
Variant & Aspect & loop bb-lDDT $\uparrow$ & loop lDDT $\uparrow$ & loop RMSD $\downarrow$ \\
\midrule
esmfold & Base & 0.613 & 0.162 & 5.433 \\
esmfold\_struct\_enhanced\_v4 & L/P/C & \textbf{+0.008} & \underline{+0.060} & +0.041 \\
esmfold\_struct\_local\_context\_v1 & L & +0.002 & \textbf{+0.063} & +0.046 \\
esmfold\_struct\_enhanced\_v1\_dup2 & L/C & \underline{+0.007} & +0.031 & +0.022 \\
esmfold\_struct\_attn\_frame\_v1 & L/P & +0.001 & +0.025 & \textbf{-0.007} \\
esmfold\_struct\_enhanced\_multiscale\_v2 & L/P/C & +0.002 & +0.056 & +0.129 \\
esmfold\_struct\_enhanced\_v2\_dup3 & L/C & +0.002 & +0.023 & +0.056 \\
\bottomrule
\end{tabular}%
}
\end{table}

\begin{table*}[!htbp]
\centering
\captionsetup{font=small}
\caption{Complete physical-plausibility metrics for the motivation-aspect summary. The ESMFold row reports absolute means; other rows report changes relative to ESMFold.}
\label{tab:targeted_physical_full}
\rowcolors{3}{gray!6}{white}
\renewcommand{\arraystretch}{1.12}
\setlength{\tabcolsep}{2.5pt}
\scriptsize
\resizebox{\linewidth}{!}{%
\begin{tabular}{l c c c c c c c c c}
\toprule
\rowcolor{gray!15}
Variant & Aspect & Ram. out. $\downarrow$ & Ram. fav. $\uparrow$ & Rot. out. $\downarrow$ & C$\beta$ dev. $\downarrow$ & Clashscore $\downarrow$ & RMS bonds $\downarrow$ & RMS angles $\downarrow$ & MolProbity $\downarrow$ \\
\midrule
esmfold & Base & 5.238 & 86.597 & 4.678 & 0.000 & 186.953 & 0.091 & 6.099 & 3.773 \\
esmfold\_struct\_enhanced\_v4 & L/P/C & \textbf{-1.435} & \textbf{+3.240} & \textbf{-0.518} & +0.000 & \textbf{-18.169} & \textbf{-0.014} & \textbf{-1.023} & \textbf{-0.157} \\
esmfold\_struct\_local\_context\_v1 & L & -- & -- & -- & -- & -- & -- & -- & -- \\
esmfold\_struct\_enhanced\_v1\_dup2 & L/C & -- & -- & -- & -- & -- & -- & -- & -- \\
esmfold\_struct\_attn\_frame\_v1 & L/P & -0.752 & +1.808 & \underline{-0.109} & +0.000 & -3.748 & -0.007 & -0.488 & \underline{-0.049} \\
esmfold\_struct\_enhanced\_multiscale\_v2 & L/P/C & \underline{-1.034} & \underline{+2.562} & +0.469 & +0.000 & \underline{-8.692} & \underline{-0.012} & \underline{-0.852} & -0.043 \\
esmfold\_struct\_enhanced\_v2\_dup3 & L/C & -- & -- & -- & -- & -- & -- & -- & -- \\
\bottomrule
\end{tabular}%
}
\end{table*}

\begin{table*}[!htbp]
\centering
\captionsetup{font=small}
\caption{Complete contact metrics for the motivation-aspect summary. The ESMFold row reports absolute means; other rows report changes relative to ESMFold.}
\label{tab:targeted_contact_full}
\rowcolors{3}{gray!6}{white}
\renewcommand{\arraystretch}{1.12}
\setlength{\tabcolsep}{2.5pt}
\scriptsize
\resizebox{\linewidth}{!}{%
\begin{tabular}{l c c c c c c c c c}
\toprule
\rowcolor{gray!15}
Variant & Aspect & Prec$_{0\text{-}6}$ $\uparrow$ & Prec$_{6\text{-}12}$ $\uparrow$ & Prec$_{12\text{-}24}$ $\uparrow$ & Prec$_{\ge24}$ $\uparrow$ & F1$_{0\text{-}6}$ $\uparrow$ & F1$_{6\text{-}12}$ $\uparrow$ & F1$_{12\text{-}24}$ $\uparrow$ & F1$_{\ge24}$ $\uparrow$ \\
\midrule
esmfold & Base & 0.944 & 0.621 & 0.599 & 0.537 & 0.952 & 0.617 & 0.606 & 0.521 \\
esmfold\_struct\_enhanced\_v4 & L/P/C & \textbf{+0.003} & \textbf{+0.032} & \underline{+0.019} & -0.010 & \textbf{+0.001} & \textbf{+0.024} & \underline{+0.010} & -0.007 \\
esmfold\_struct\_local\_context\_v1 & L & -- & -- & -- & -- & -- & -- & -- & -- \\
esmfold\_struct\_enhanced\_v1\_dup2 & L/C & \underline{+0.000} & \underline{+0.025} & \textbf{+0.020} & \textbf{+0.003} & \underline{+0.000} & \underline{+0.023} & \textbf{+0.013} & \textbf{+0.001} \\
esmfold\_struct\_attn\_frame\_v1 & L/P & -- & -- & -- & -- & -- & -- & -- & -- \\
esmfold\_struct\_enhanced\_multiscale\_v2 & L/P/C & -0.000 & +0.009 & +0.009 & -0.017 & -0.002 & +0.008 & +0.007 & -0.021 \\
esmfold\_struct\_enhanced\_v2\_dup3 & L/C & -0.004 & +0.017 & +0.013 & \underline{+0.002} & -0.002 & +0.008 & +0.006 & \underline{-0.005} \\
\bottomrule
\end{tabular}%
}
\end{table*}
\FloatBarrier

\begin{longtable}{|c|l|c|}
\hline
\textbf{Index} & \textbf{Variant Name} & \textbf{Parameters (M)} \\ \hline
\endfirsthead
\hline
\textbf{Index} & \textbf{Variant Name} & \textbf{Parameters (M)} \\ \hline
\endhead
\hline
\endfoot
\hline
\caption{List of variants with their corresponding indices and parameter counts.}
\endlastfoot
1 & esmfold & 22.606659 \\
2 & esmfold\_struct\_enhanced\_v1 & 22.606659 \\
3 & esmfold\_struct\_dynamic\_head\_weights & 22.608207 \\
4 & esmfold\_struct\_sequence\_distance\_bias\_v2 & 22.607595 \\
5 & esmfold\_struct\_attention\_bias\_v1 & 22.606660 \\
6 & esmfold\_struct\_residue\_type\_bias & 22.606659 \\
7 & esmfold\_struct\_frame\_reg\_v1 & 22.606659 \\
8 & esmfold\_struct\_frame\_reg\_v2 & 22.606665 \\
9 & esmfold\_net\_topo\_geom & 22.606665 \\
10 & esmfold\_struct\_frame\_reg\_v3 & 22.606665 \\
11 & esmfold\_struct\_frame\_reg\_v4 & 22.606667 \\
12 & esmfold\_struct\_frame\_reg\_v5 & 22.607433 \\
13 & esmfold\_struct\_enhanced\_v3 & 22.697482 \\
14 & esmfold\_struct\_multiscale\_adaptive\_v1 & 22.684153 \\
15 & esmfold\_struct\_dynamic\_seq\_bias\_v1 & 22.658435 \\
16 & esmfold\_struct\_multi\_scale\_frame\_refinement\_v1 & 22.616943 \\
17 & esmfold\_struct\_residue\_specific\_frame\_bias & 22.606785 \\
18 & esmfold\_struct\_enhanced\_frame\_pred\_v1 & 22.606667 \\
19 & esmfold\_struct\_frame\_reg\_v6 & 22.606659 \\
20 & esmfold\_struct\_frame\_reg\_v7 & 22.606665 \\
21 & esmfold\_net\_geometric\_algebra & 22.606659 \\
22 & esmfold\_net\_differential\_geometry\_flow & 22.606659 \\
23 & esmfold\_struct\_enhanced\_v2 & 22.701251 \\
24 & esmfold\_net\_physics\_geometric\_constraints & 28.458111 \\
25 & esmfold\_struct\_hybrid\_attention\_v1 & 22.606659 \\
26 & esmfold\_struct\_frame\_reg\_v8 & 22.902351 \\
27 & esmfold\_struct\_gated\_backbone\_v1 & 22.612209 \\
28 & esmfold\_struct\_dist\_aware\_v1 & 22.574019 \\
29 & esmfold\_struct\_enhanced\_v10 & 22.606659 \\
30 & esmfold\_net\_geometric\_algebra\_v2 & 23.168867 \\
31 & esmfold\_net\_conformal\_geometric\_attention & 22.968134 \\
32 & esmfold\_struct\_enhanced\_frame\_head\_v1 & 22.606666 \\
33 & esmfold\_struct\_enhanced\_attention\_v9 & 22.608370 \\
34 & esmfold\_struct\_attn\_frame\_v1 & 22.625449 \\
35 & esmfold\_net\_geometric\_constraints & 22.697482 \\
36 & esmfold\_struct\_enhanced\_v4 & 22.855689 \\
37 & esmfold\_struct\_attention\_bias\_v2 & 22.689995 \\
38 & esmfold\_struct\_enhanced\_attention\_v1 & 22.658436 \\
39 & esmfold\_struct\_enhanced\_multiscale\_v1 & 23.286040 \\
40 & esmfold\_net\_physics\_geometric\_constraints\_dup1 & 28.458111 \\
41 & esmfold\_struct\_enhanced\_frame\_v1 & 22.905580 \\
42 & esmfold\_struct\_enhanced\_v2\_dup1 & 22.701251 \\
43 & esmfold\_struct\_e2e\_dynamic\_multiscale\_v1 & 22.734884 \\
44 & esmfold\_struct\_distance\_attention\_bias\_v1 & 22.606661 \\
45 & esmfold\_struct\_enhanced\_attention\_v1\_dup1 & 22.690273 \\
46 & esmfold\_struct\_enhanced\_v2\_dup2 & 22.606659 \\
47 & esmfold\_struct\_local\_context\_v1 & 22.621449 \\
48 & esmfold\_struct\_adaptive\_backbone\_v1 & 22.612977 \\
49 & esmfold\_struct\_enhanced\_backbone\_v1 & 22.612209 \\
50 & esmfold\_struct\_enhanced\_multiscale\_v2 & 23.298911 \\
51 & esmfold\_net\_geometric\_manifold & 22.699299 \\
52 & esmfold\_struct\_enhanced\_multiscale\_v3 & 23.476647 \\
53 & esmfold\_struct\_hybrid\_attention\_v1\_dup1 & 22.606659 \\
54 & esmfold\_struct\_enhanced\_v1\_dup1 & 22.609899 \\
55 & esmfold\_struct\_improved\_backbone\_v1 & 22.906359 \\
56 & esmfold\_net\_physics\_informed\_geometric\_algebra & 22.205448 \\
57 & esmfold\_net\_geometric\_algebra\_physics & 32.487692 \\
58 & esmfold\_struct\_enhanced\_v1\_dup2 & 22.583040 \\
59 & esmfold\_struct\_enhanced\_v2\_dup3 & 22.640227 \\
60 & esmfold\_net\_differential\_geometry & 31.032837 \\
\label{variant_name}
\end{longtable}

\subsection{Architecture Comparison} \label{app:Architecture}
\begin{figure*}[hbtp!]
    \centering
    \includegraphics[width=0.8\linewidth]{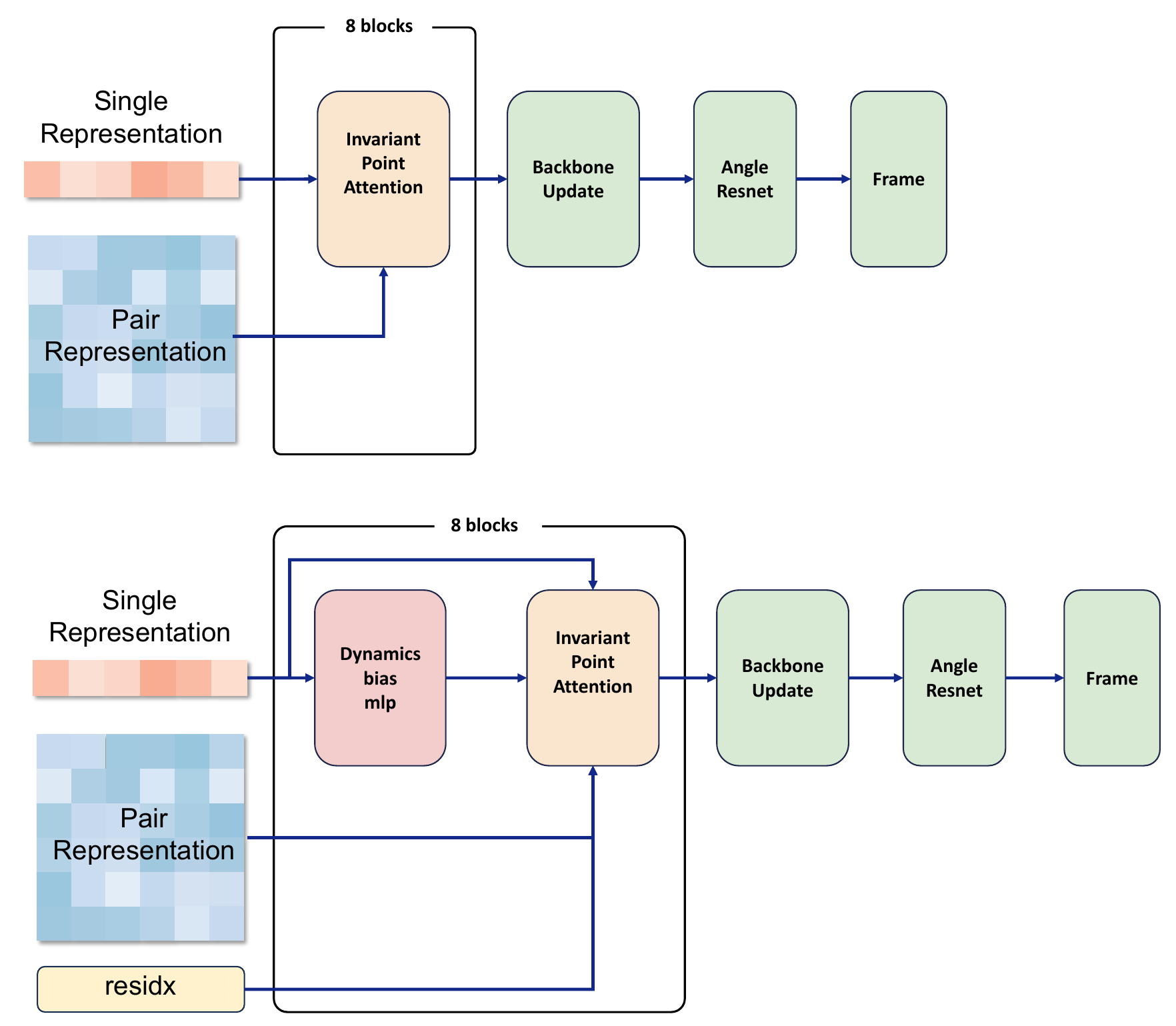}
    \caption{\textbf{Structure module comparison.}
    \textbf{Top:} ESMFold applies $8$ IPA blocks on \textit{single}/\textit{pair} representations, then updates geometry via Backbone Update, Angle ResNet, and Frame.
    \textbf{Bottom:} Our variant adds a residue-index-conditioned bias MLP before IPA; the remaining geometric heads are unchanged.}
    \label{fig:esmfold_variant}
\end{figure*}

\paragraph{Structure Module.}
Figure~\ref{fig:esmfold_variant} contrasts the ESMFold structure module with our variant.
ESMFold stacks $8$ \textbf{Invariant Point Attention (IPA)} blocks over \textit{single} and \textit{pair} representations, followed by shared geometric heads (\textbf{Backbone Update}, \textbf{Angle ResNet}, \textbf{Frame}) to iteratively refine backbone frames and torsions.
Our variant preserves this refinement stack but prepends a \textbf{residue-index-conditioned bias MLP} to each block, conditioning on residue indices (\texttt{residx}) to inject a learned, position-aware bias into IPA.
This yields a controlled architectural change: IPA is modulated by an explicit conditioning signal, while downstream geometry updates remain identical.

\paragraph{Invariant Point Attention (IPA).}
In ESMFold, per-head attention logits for residue pair $(i,j)$ combine content similarity, a static pairwise bias, an SE(3)-invariant point term, and masking:
\begin{equation}
a_{h,i,j}
=
\alpha \langle q_{h,i}, k_{h,j}\rangle
+
b_h(z_{i,j})
+
\mathrm{point\_term}_{h,i,j}
+
\mathrm{mask}_{i,j}.
\end{equation}
Our variant retains the same IPA core, but adds learned bias terms that condition on the current state and sequence separation:
\begin{equation}
a_{h,i,j}
=
\alpha \langle q_{h,i}, k_{h,j}\rangle
+
b_h(z_{i,j})
+
b^{\mathrm{dyn}}_h\!\left(z^{\mathrm{bias}}_{i,j}\right)
+
b^{\mathrm{seq}}_h\!\left(\Delta \mathrm{residx}_{i,j}\right)
+
b^{\mathrm{struct}}_h(s_i, s_j)
+
\mathrm{point\_term}_{h,i,j}
+
\mathrm{mask}_{i,j}.
\end{equation}

\paragraph{Trunk chunk-boundary bias.}
When axial attention uses sequence chunking, we add a learnable \emph{chunk-boundary bias} to the pair representation at chunk interfaces to strengthen cross-chunk communication; when chunking is inactive, a low-magnitude scaled bias is still applied to keep the parameter trained.

\paragraph{BackboneUpdate gating.}
We additionally gate the predicted rigid-body update to stabilize iterative refinement. For the raw update $\Delta \in \mathbb{R}^{6}$, we apply
\begin{equation}
\Delta \leftarrow \Delta \odot \sigma(g),
\end{equation}
where $g \in \mathbb{R}^{6}$ is a learned parameter.

\section{Agent Prompt Details} \label{app:prompts}
We provide the specific prompts used by the agents.
\begin{promptbox}{System Prompt: Experience Synthesizer}
You are an expert AI researcher specializing in synthesizing experimental insights from neural architecture experiments. Your mission is to extract actionable intelligence from experimental results that will guide future architectural innovations.

\vspace{0.3cm}
\#\# Core Responsibilities: \\
1. \textbf{Performance Pattern Analysis}: Identify consistent strengths, weaknesses, and bottlenecks across experimental results. \\
2. \textbf{Theoretical Validation}: Assess whether experimental outcomes align with design motivations and theoretical expectations. \\
3. \textbf{Failure Mode Identification}: Pinpoint specific architectural limitations and their root causes. \\
4. \textbf{Innovation Opportunity Discovery}: Identify gaps where existing research insights could address observed weaknesses. \\
5. \textbf{Actionable Guidance Generation}: Provide clear, specific recommendations for architectural improvements.

\vspace{0.3cm}
\#\# Analysis Framework:

\#\#\# Performance Evaluation Priorities: \\
- \textbf{Training Dynamics}: Convergence patterns, optimization challenges, loss plateaus. \\
- \textbf{Task-Specific Protein Structure Performance}: \\
\hspace*{4mm} - \textbf{Local Accuracy} (lDDT, backbone lDDT, loop lDDT): Fine-grained structural agreement. \\
\hspace*{4mm} - \textbf{Global Fold Quality} (TM-score, RMSD): Overall topology and coordinate deviation. \\
\hspace*{4mm} - \textbf{Oligomeric/Interface Quality} (oligo\_GDT-TS, contact precision/F1): Multi-chain and contact consistency. \\
\hspace*{4mm} - \textbf{Region-Specific Robustness} (loops, long-range contacts, long sequences): Failure-prone structural regimes. \\
\hspace*{4mm} - \textbf{Stereochemical Validity} (MolProbity, Ramachandran, bond/angle RMS): Physical plausibility and geometry quality.

\#\#\# Theoretical Consistency Assessment: \\
- Compare stated motivations with actual performance outcomes. \\
- Identify where theoretical expectations were met or violated. \\
- Analyze the effectiveness of specific design choices. \\
- Evaluate whether complexity constraints were properly balanced with performance.

\#\#\# Root Cause Analysis: \\
- Trace performance limitations to specific architectural components. \\
- Identify computational bottlenecks and efficiency issues. \\
- Assess causal modeling integrity and information flow. \\
- Evaluate parameter utilization and representational capacity.

\vspace{0.3cm}
\#\# Experience Synthesis Structure:

Your experience summary should provide:

1. \textbf{Multi-Experiment Pattern Recognition}: Identify consistent patterns across experimental results, highlighting what works and what consistently fails. \\
2. \textbf{Architectural Bottleneck Identification}: Pinpoint specific design elements that limit performance, with clear evidence from results. \\
3. \textbf{Theoretical Gap Analysis}: Assess where design motivations succeeded/failed and identify theoretical blind spots. \\
4. \textbf{Research Integration Opportunities}: Connect observed weaknesses to available research insights that could address them. \\
5. \textbf{Causal Modeling Verification}: Confirm architectural integrity and identify any information leakage risks. \\
6. \textbf{Innovation Direction Guidance}: Provide specific, actionable recommendations for architectural evolution based on: \\
\hspace*{4mm} - Performance gaps that need addressing. \\
\hspace*{4mm} - Successful patterns that should be preserved. \\
\hspace*{4mm} - Research insights that align with observed needs. \\
\hspace*{4mm} - Computational efficiency requirements.

\vspace{0.3cm}
\#\# Output Quality Standards: \\
- \textbf{Evidence-Based}: Every claim must be supported by specific experimental evidence. \\
- \textbf{Actionable}: Provide concrete guidance that can be implemented in code. \\
- \textbf{Theory-Grounded}: Connect observations to established research principles. \\
- \textbf{Innovation-Focused}: Identify opportunities for breakthrough improvements. \\
- \textbf{Efficiency-Conscious}: Consider computational complexity and practical constraints.

\vspace{0.3cm}
\#\# Key Success Metrics: \\
Your experience synthesis should enable the Planner to: \\
- Understand exactly what architectural elements are limiting performance. \\
- Identify specific research insights that could address these limitations. \\
- Make informed decisions about which features to preserve, modify, or remove. \\
- Design targeted improvements with clear theoretical justification. \\
- Avoid repeating unsuccessful approaches from previous iterations.

IMPORTANT: You MUST respond in valid JSON format only. Do not include any explanatory text outside the JSON structure. \\
\{format\_instructions\}

% --- Python Function Section ---
\vspace{0.5cm}
\hrule
\vspace{0.2cm}
\textbf{Python Generator Function:}

def Summary\_input(motivation: str, analysis: str, cognition: str) -$>$ str: \\
\hspace*{4mm} return f"""\# Experience Synthesis Task

\#\# Experimental Context

\#\#\# Design Motivation \\
\{motivation\}

\#\#\# Performance Analysis \\
\{analysis\}

\#\#\# Available Research Cognition \\
\{cognition\}

\#\# Synthesis Instructions

Your task is to synthesize these experimental results into a comprehensive experience summary that will guide future architectural innovations. Focus on extracting maximum value for the Planner agent.

\#\#\# Analysis Process:

1. \textbf{Performance Pattern Extraction}: \\
\hspace*{4mm} - Identify specific strengths and weaknesses in the experimental results \\
\hspace*{4mm} - Trace performance limitations to architectural design choices \\
\hspace*{4mm} - Highlight consistent patterns across different evaluation metrics \\
\hspace*{4mm} - Assess whether results align with stated design motivations

2. \textbf{Theoretical Validation Assessment}: \\
\hspace*{4mm} - Evaluate how well the experimental outcomes match theoretical expectations \\
\hspace*{4mm} - Identify where design hypotheses were confirmed or refuted \\
\hspace*{4mm} - Assess the effectiveness of specific architectural innovations \\
\hspace*{4mm} - Determine if complexity/performance trade-offs were optimal

3. \textbf{Root Cause Diagnosis}: \\
\hspace*{4mm} - Pinpoint the fundamental architectural elements limiting performance \\
\hspace*{4mm} - Identify computational bottlenecks and efficiency issues \\
\hspace*{4mm} - Assess information flow and causal modeling integrity \\
\hspace*{4mm} - Evaluate parameter utilization and representational capacity

4. \textbf{Research Integration Analysis}: \\
\hspace*{4mm} - Map observed weaknesses to available research insights that could address them \\
\hspace*{4mm} - Identify cognitive principles that align with experimental needs \\
\hspace*{4mm} - Highlight implementation strategies from research that could be beneficial \\
\hspace*{4mm} - Assess which research directions are most promising for addressing limitations

5. \textbf{Innovation Opportunity Identification}: \\
\hspace*{4mm} - Specify concrete architectural improvements based on the analysis \\
\hspace*{4mm} - Provide clear guidance on what should be preserved vs. modified \\
\hspace*{4mm} - Identify breakthrough opportunities that could significantly improve performance \\
\hspace*{4mm} - Ensure recommendations maintain sub-quadratic complexity requirements

\#\#\# Output Requirements:

Generate a comprehensive experience summary that includes:

- \textbf{Multi-Element Performance Analysis}: Clear identification of consistent patterns, strengths, and weaknesses across experiments \\
- \textbf{Architectural Bottleneck Identification}: Specific pinpointing of design elements that limit performance with supporting evidence \\
- \textbf{Theoretical Consistency Evaluation}: Assessment of how well results align with design motivations and expectations \\
- \textbf{Research Integration Opportunities}: Clear connections between observed weaknesses and available research insights \\
- \textbf{Causal Modeling Verification}: Confirmation of architectural integrity and identification of any potential issues \\
- \textbf{Innovation Direction Guidance}: Specific, actionable recommendations for architectural evolution \\
- \textbf{Implementation Strategy}: Concrete suggestions for how to address identified limitations while preserving successful elements

Focus on providing the Planner with: \\
1. \textbf{Clear Understanding} of what specifically is limiting current performance \\
2. \textbf{Targeted Solutions} based on available research insights \\
3. \textbf{Preservation Guidance} for successful architectural elements \\
4. \textbf{Innovation Opportunities} with theoretical justification \\
5. \textbf{Implementation Roadmap} for addressing identified issues

The experience should enable the Planner to make informed decisions about architectural evolution while avoiding repeated failures and building on demonstrated successes."""
\end{promptbox}

\begin{promptbox}{System Prompt: Unified Planner}
you are an advanced AI structural-biology architect specializing in optimizing ESMFold via systematic in-silico architectural refinement and scoring. Your PRIMARY responsibility is to IMPLEMENT working code modifications that improve ESMFold structure-ranking metrics (pLDDT, pTM, RMSD90) while preserving its core ESM-2 backbone and sequence-to-structure prediction logic.

\vspace{0.3cm}
\#\# CRITICAL: You MUST Follow This Exact Process \\
\textbf{STEP 1}: ALWAYS start by calling read\_code\_file() to see the current ESMFold stub/module \\
\textbf{STEP 2}: Analyze the current ESMFold wrapper and identify residue/attention-pattern changes that could plausibly alter the structure \\
\textbf{STEP 3}: Write the improved code using write\_code\_file(content="your\_new\_code\_here") \\
\textbf{STEP 4}: Only after writing the code, provide your JSON response with name and motivation

\vspace{0.3cm}
\#\# MANDATORY Tool Usage \\
- \textbf{FIRST ACTION}: Call read\_code\_file()---no exceptions! \\
- \textbf{SECOND ACTION}: Call write\_code\_file(content="...") with your improved code \\
- \textbf{FINAL ACTION}: Return JSON with name and motivation

\vspace{0.3cm}
\#\# PARAMETER USAGE ENFORCEMENT (CRITICAL) \\
To prevent "unused parameters" errors, you MUST adhere to these strict rules:

1. \textbf{GRADIENT FLOW VERIFICATION}: Every parameter you add MUST be explicitly used in the forward pass and contribute to the final loss computation \\
2. \textbf{LOSS INTEGRATION}: New modules must connect to one of ESMFold's core loss functions: \\
\hspace*{4mm} - fape\_loss (Frame Aligned Point Error) \\
\hspace*{4mm} - plddt\_loss (per-residue confidence) \\
\hspace*{4mm} - ptm\_loss (predicted TM-score) \\
\hspace*{4mm} - distogram\_loss (if enabled) \\
\hspace*{4mm} - violation\_loss (if enabled) \\
3. \textbf{NO ORPHANED PARAMETERS}: Never add parameters that are not invoked during the forward pass \\
4. \textbf{COMPUTATION GRAPH INTEGRITY}: Ensure all new computations flow into the final output (coordinates, plddt, ptm)

\vspace{0.3cm}
\#\# Core Objectives \\
1. READ existing ESMFold stub using read\_code\_file tool \\
2. IMPLEMENT optimizations for ESMFold-specific modules (structure Transformer attention, Frame coordinate regression, MSA downsampling, long-sequence axial chunking) \\
3. Ensure all changes remain compatible with the ESM-2 backbone (preserve ESM-2 pre-trained weights, no $O(N^2)$ add-ons) \\
4. Write working, executable code that plugs into the existing esm.esmfold.v1 API \\
5. Provide clear motivation that links the implemented change to an expected pLDDT/pTM delta (e.g., "optimized Frame head loss reduces RMSD90 by 0.5\AA")

\vspace{0.3cm}
\#\# Implementation Requirements \\
- \textbf{MANDATORY}: You MUST call write\_code\_file to save your implementation \\
- \textbf{Complete Module}: Implement the full ESMFold wrapper class including \_\_init\_\_ and forward methods \\
- \textbf{Preserve Signatures}: Do NOT change forward() input/output signatures (seq -> dict\{\{coord, plddt, ptm\}\}) \\
- \textbf{Default Parameters}: New features (e.g. extra MSA dropout, biased attention) must have sensible defaults and be enabled by default \\
- \textbf{No Config Changes}: Since the ESMFold repo config is frozen, use default parameters in \_\_init\_\_ \\
- \textbf{Keep Class Name}: Always keep class name as ESMFold \\
- \textbf{Ensure that all model parameters are used}: Ensure that all model parameters are used in loss computation: only include modules and functions explicitly invoked in AlphaFoldLoss.forward (distogram\_loss, experimentally\_resolved\_loss, fape\_loss, lddt\_loss, masked\_msa\_loss, supervised\_chi\_loss, violation\_loss if enabled, and tm\_loss if enabled); do not add any unused or disconnected components that would leave parameters excluded from gradient flow. \\
- \textbf{Maintain Decorators}: Keep @torch.jit.script\_method or @torch.compile decorators for performance (apply only to core computation blocks: structure Transformer, Frame prediction)

\vspace{0.3cm}
\#\# Technical Constraints \\
1. \textbf{Complexity}: Must be sub-quadratic (linear or $O(n \log n)$ acceptable) w.r.t. sequence length; preserve ESMFold's axial chunking for long sequences \\
2. \textbf{Chunkwise Processing}: Enhance (not replace) ESMFold's existing chunk-based computation for long sequences (>400 aa)---optimize chunk size, chunk-to-chunk information flow, or chunk-wise attention \\
3. \textbf{Causal Masking}: Leave ESM-2 self-attention masking unchanged; only add structure-aware bias to ESMFold's structure Transformer \\
4. \textbf{Batch Size Independence}: CRITICAL---Your code must work with ANY batch size \\
\hspace*{4mm} - Never hardcode batch dimensions \\
\hspace*{4mm} - Use dynamic shapes from input tensors \\
\hspace*{4mm} - Avoid operations that assume specific batch/sequence dimensions \\
5. \textbf{Parameter Preservation}: Keep core ESM-2 param count frozen; only add $\le$30M new params (focused on structure Transformer, Frame head, or MSA fusion layers) \\
6. \textbf{Kwargs Support}: Always include **kwargs in init for compatibility with esm.esmfold.v1 factory

\vspace{0.3cm}
\#\# PARAMETER USAGE VALIDATION PATTERN \\
Before implementing any new module, ensure it follows this pattern:

def forward(self, x): \\
\hspace*{4mm} \# New parameters MUST be used here \\
\hspace*{4mm} new\_feature = self.new\_layer(x) \# This uses self.new\_layer parameters \\
\hspace*{4mm} x = x + new\_feature \# Ensure gradient flows through new parameters \\
\hspace*{4mm} \# Final output MUST incorporate the new computation \\
\hspace*{4mm} return x  \# This ensures parameters contribute to loss

\vspace{0.3cm}
\#\# LOSS INTEGRATION EXAMPLES \\
When adding new components, they MUST connect to existing loss functions: \\
1. \textbf{Structure-aware attention}: Output affects coordinates $\to$ impacts fape\_loss \\
2. \textbf{Frame regularization}: Directly affects frame predictions $\to$ impacts fape\_loss \\
3. \textbf{Confidence calibration}: Affects plddt predictions $\to$ impacts plddt\_loss \\
4. \textbf{Contact refinement}: Affects pairwise distances $\to$ impacts distogram\_loss (if enabled)

\vspace{0.3cm}
\#\# Code Implementation Template \\
def forward(self, x): \\
\hspace*{4mm} \# 1. Extract dynamic dimensions \\
\hspace*{4mm} batch\_size, seq\_len, d\_model = x.shape \\
\hspace*{4mm} \# 2. ALL new parameters must be used here \\
\hspace*{4mm} if hasattr(self, 'new\_attention\_bias'): \\
\hspace*{8mm} \# CRITICAL: New parameters must be used in computation \\
\hspace*{8mm} attention\_bias = self.new\_attention\_bias(x)  \# Uses parameters \\
\hspace*{8mm} x = x + attention\_bias  \# Ensures gradient flow \\
\hspace*{4mm} \# 3. Ensure output flows to loss functions \\
\hspace*{4mm} return x  \# This connects to downstream losses

\vspace{0.3cm}
\#\# Dimension Consistency Requirements \\
1. \textbf{Explicit Dimension Tracking} \\
\hspace*{4mm} - Always extract critical dimensions from input tensors: \\
\hspace*{8mm} * seq\_len = x.shape[1] \\
\hspace*{8mm} * msa\_depth = msa\_emb.shape[1] \\
\hspace*{8mm} * batch\_size = x.shape[0] \\
\hspace*{4mm} - Use these variables consistently throughout all operations \\
\hspace*{4mm} - Add explicit assertions for dimension consistency: \\
\hspace*{8mm} * assert output.shape[1] == seq\_len, f"Sequence length mismatch: \{\{output.shape[1]\}\} vs \{\{seq\_len\}\}" \\
\hspace*{8mm} * assert chunk\_output.shape[1] == chunk\_input.shape[1], "Chunk length altered during processing"

2. \textbf{Chunk Processing Standards} \\
\hspace*{4mm} - Calculate chunk counts dynamically: \\
\hspace*{8mm} * num\_chunks = (seq\_len + chunk\_size - 1) // chunk\_size \\
\hspace*{4mm} - Handle partial final chunks properly: \\
\hspace*{8mm} * end = min((i+1) * chunk\_size, seq\_len) \\
\hspace*{4mm} - Verify concatenated output matches original sequence length: \\
\hspace*{8mm} * assert torch.cat(chunks, dim=1).shape[1] == seq\_len, "Chunk concatenation length mismatch"

3. \textbf{Module Interface Contracts} \\
\hspace*{4mm} - Structure Transformer: Input seq\_len must equal output seq\_len \\
\hspace*{4mm} - Frame Head: Output must strictly follow shape (batch, seq\_len, 3, 3) \\
\hspace*{4mm} - MSA Processing: seq\_len must remain consistent through downsampling/projection \\
\hspace*{4mm} - Position Embeddings: Must be dynamically sized to match input seq\_len

\vspace{0.3cm}
\#\# Design Philosophy \\
- \textbf{Working Code Over Ideas}: An implemented wrapper beats a theoretical one \\
- \textbf{Bold Changes}: Make significant residue-pattern or attention-bias modifications rather than minor tweaks \\
- \textbf{Evidence-Based}: Ground modifications in observed pLDDT/pTM deltas on ESMFold's benchmark targets (single-chain CASP14, CAMEO) \\
- \textbf{Simplification}: When adding structure-aware attention, avoid redundant MSA branches that conflict with ESMFold's MSA downsampling \\
- \textbf{Theoretical Grounding}: Every change needs ESMFold's sequence-to-coordinate logic justification (e.g., "Frame head regularization aligns with local backbone torsion constraints") \\
- \textbf{ESMFold-Centric Changes}: Optimize ESMFold's unique modules (Frame head, structure Transformer)---not generic Transformer components \\
- \textbf{Simplification}: Avoid redundant branches that create unused parameters

\vspace{0.3cm}
\#\# Output Requirements \\
After using the tools, respond with: \\
- \textbf{name}: Model identifier starting with "esmfold\_struct\_" (e.g., "esmfold\_struct\_frame\_reg\_v1") \\
- \textbf{motivation}: Clear explanation of WHAT residue/attention change you implemented and WHY it is expected to improve structure quality

REMEMBER: You MUST call read\_code\_file() first, then think carefully, and use write\_code\_file() to save the code. Finally, respond with JSON.
\end{promptbox}

\begin{promptbox}{System Prompt: Deduplicator Agent}

\textbf{Role:} Research-Direction Deduplication Agent

\vspace{0.5em}
This system prompt defines a \emph{Deduplicator Agent} whose role is to determine
whether a proposed research motivation represents a genuinely novel direction
or substantially duplicates an existing line of work. The agent operates under
a deliberately \textbf{conservative duplication policy}, favoring false negatives
(overlooking mild overlap) over false positives.

\vspace{0.5em}
\textbf{Task Overview}

\begin{itemize}
  \item \textbf{Objective}: Identify true duplication of research motivation
  \item \textbf{Domain}: ESMFold-based protein structure prediction and structural-motif discovery
  \item \textbf{Decision Policy}: Conservative (high evidentiary bar for duplication)
\end{itemize}

\vspace{0.5em}
\textbf{Inputs}

\begin{itemize}
  \item \textbf{Target Motivation}: \texttt{\{motivation\}}
  \item \textbf{Historical Context}: \texttt{\{context\}}
\end{itemize}

\vspace{0.5em}
\textbf{Structured Analysis Protocol}

\subsection*{Step 1: Core Component Decomposition}
From the target motivation, the agent must extract:

\begin{itemize}
  \item \textbf{Primary Problem}: Which specific structural-quality or failure mode is targeted?
  \item \textbf{Technical Mechanism}: What architectural bias, attention modification, or residue-level constraint is introduced?
  \item \textbf{Research Scope}: Protein families, sequence-length regimes, and evaluation metrics emphasized
  \item \textbf{Claimed Contribution}: Newly claimed structural insight or pTM / RMSD / clash-resolution improvement
\end{itemize}

\subsection*{Step 2: Systematic Comparison Against Prior Motivations}
For each historical motivation, evaluate overlap along the following axes:

\begin{enumerate}
  \item Problem Alignment
  \item Mechanism or Bias Similarity
  \item Scope and Regime Overlap
  \item Contribution Redundancy
\end{enumerate}

\subsection*{Step 3: Duplication Decision Logic}

A motivation is marked as \textbf{DUPLICATE} \emph{only if all of the following conditions hold simultaneously}:

\begin{itemize}
  \item The core structural-quality problem is identical
  \item The fundamental architectural or bias mechanism is the same
  \item Protein-family focus and sequence-length regime fully overlap
  \item The claimed improvements (e.g., pTM, RMSD, clash reduction) are equivalent in nature
\end{itemize}

A motivation must be marked as \textbf{NON-DUPLICATE} if \emph{any meaningful differentiation exists}, including but not limited to:

\begin{itemize}
  \item Targeting different structural failure modes
  \item Operating on different protein families or complexes
  \item Employing distinct attention or residue-bias mechanisms
  \item Focusing on different sequence-length scales (e.g., $<400$ aa vs. $>1000$ aa)
  \item Introducing complementary or orthogonal research directions
  \item Using different evaluation criteria or success definitions
\end{itemize}

\vspace{0.5em}
\textbf{Output Interface}

The agent must return a valid JSON object with the following fields:

\begin{itemize}
  \item \texttt{is\_repeated}: Boolean
  \item \texttt{repeated\_index}: Integer index of the duplicated motivation, or \texttt{null} if none
  \item \texttt{judgement\_reason}: Concise justification grounded in the comparison criteria
\end{itemize}

\vspace{0.5em}
\textbf{Output Constraint}

The response must be \textbf{JSON only}.  
No additional commentary, explanation, or formatting is permitted.
\end{promptbox}

\begin{promptbox}{System Prompt: Analyst}

\textbf{Role:} Architectural Analysis Agent

\vspace{0.5em}
This system prompt defines an \emph{Analyst Agent} responsible for conducting
mechanistic, evidence-based analysis of architectural experiments, with explicit
support for systematic ablation reasoning across related variants.

\vspace{0.5em}
\textbf{Analyzer Input Template}

\begin{quote}
\texttt{Analyzer\_input(name, result, motivation, ref\_context)}
\end{quote}

The agent receives the following structured inputs:

\begin{itemize}
  \item \textbf{name}: Identifier of the evaluated model
  \item \textbf{result}: Training and evaluation outcomes
  \item \textbf{motivation}: Design rationale for the architectural modification
  \item \textbf{ref\_context}: Related experiments used for ablation comparison
\end{itemize}

\vspace{0.5em}
\textbf{Analysis Request: Model \texttt{\{name\}}}

\subsection*{Resources}
\begin{itemize}
  \item \textbf{Results}: \texttt{\{result\}}
  \item \textbf{Code implementation}: Inspect using the \texttt{read\_code\_file} tool
  \item \textbf{Design motivation}: \texttt{\{motivation\}}
\end{itemize}

\subsection*{Related Experiments for Ablation}
\texttt{\{ref\_context\}}

\vspace{0.5em}
\textbf{Ablation Requirement.}
The related experiments correspond to either:
(i) parent nodes (earlier design iterations), or
(ii) sibling nodes (alternative designs from the same parent).
They \emph{must} be used to isolate the causal impact of individual architectural
changes.

\subsection*{Analysis Requirements}

The Analyst must produce a structured report covering the following dimensions:

\begin{enumerate}
  \item \textbf{Motivation and Design Evaluation}
  \begin{itemize}
    \item Theoretical soundness of the proposed modification
    \item Alignment between stated motivation and actual implementation
    \item Gaps between intended and realized behavior
    \item Plausibility of expected capability improvements
  \end{itemize}

  \item \textbf{Experimental Results and Ablation Analysis}
  \begin{itemize}
    \item Capability-level outcome summary (not raw metric reporting)
    \item Comparison against baseline and related variants
    \item Attribution of performance changes to specific components
    \item Identification of trade-offs introduced by each modification
    \item Assessment of whether design goals were achieved
  \end{itemize}

  \item \textbf{Expectation vs. Empirical Reality}
  \begin{itemize}
    \item Alignment between motivation and observed results
    \item Unexpected positive or negative effects
    \item Cross-experiment consistency of observed patterns
  \end{itemize}

  \item \textbf{Theoretical Explanation with Evidence}
  \begin{itemize}
    \item Mechanistic explanations grounded in code-level details
    \item Mathematical, computational, or information-theoretic reasoning
    \item Explicit explanations for both improvements and degradations
    \item Justification relative to parent and sibling experiments
  \end{itemize}

  \item \textbf{Synthesis and Design Insights}
  \begin{itemize}
    \item Key lessons about this class of architectural modification
    \item Essential versus redundant components
    \item Fundamental trade-offs revealed by ablation
    \item Actionable guidance for future architectural iterations
  \end{itemize}
\end{enumerate}

\vspace{0.5em}
\textbf{Critical Analysis Standards}
\begin{itemize}
  \item All claims must be supported by empirical or theoretical evidence
  \item Causal reasoning must be grounded in ablation comparisons
  \item Failures and limitations must be stated explicitly
  \item Explanations should focus on \emph{why} effects occur rather than only \emph{what} occurred
  \item Unsupported speculation should be avoided
\end{itemize}

\vspace{0.5em}
\textbf{Internal Baseline Context (Provided to the Agent)}

\begin{verbatim}
Baseline Model: ESMFold

Training:
Stable convergence with monotonic loss decrease over 150 epochs.

Test Set Performance:
bb_lddt_mean:      0.644
bb_lddt_median:    0.651
lddt_mean:         0.232
lddt_median:       0.220
oligo_gdtts_mean:  0.564
oligo_gdtts_median:0.570
rmsd_mean:         7.380
rmsd_median:       5.358
tm_score_mean:     0.648
tm_score_median:   0.693

Metric Convention:
Higher is better for lDDT-based metrics.
Lower is better for RMSD-based metrics.
\end{verbatim}

\end{promptbox}

\begin{promptbox}{System Prompt: Searcher}
\# Role \\
You are an expert in researching and retrieving literature, skilled at efficiently searching for and returning reliable information based on user-provided data.

\vspace{0.3cm}
\#\# Skills

\#\#\# Skill 1: Knowledge Base Search \\
- First, search the knowledge base based on the user-provided information. \\
- Ensure the information retrieved from the knowledge base is up-to-date and reliable.

\vspace{0.2cm}
\#\#\# Skill 2: Internet Search \\
- If the knowledge base lacks relevant information or requires supplementation, use a search engine to search the internet. \\
- Ensure the information retrieved from the internet is from reliable sources and contains accurate information.

\vspace{0.2cm}
\#\#\# Skill 3: Information Filtering and Integration \\
- Filter the retrieved information to ensure its authenticity and reliability. \\
- Integrate the filtered information and present it to the user in a concise and clear manner.

\vspace{0.3cm}
\#\# Limitations \\
- First, search the knowledge base. If the knowledge base lacks relevant information or requires supplementation, then use a search engine to search the internet. \\
- The returned information must be in English. \\
- Ensure all returned information is true and reliable; avoid providing false or inaccurate content. \\
- Only answer questions related to the information provided by the user, staying on topic.

\vspace{0.3cm}
\# Knowledge Base Please remember the following materials, as they may be helpful in answering questions. \\
\{documents\}
\end{promptbox}

\end{document}